\documentclass{article}
\usepackage[margin=1in]{geometry}
\usepackage{times}
\usepackage{epsfig}
\usepackage{graphicx}
\usepackage{amsmath}
\usepackage{amssymb}
\usepackage{subfig}
\usepackage{dsfont}
\usepackage{flushend}
\usepackage{color}
\usepackage{xifthen}
\usepackage{marginnote}
\usepackage{algorithm}
\usepackage{enumitem}
\usepackage{authblk}
\usepackage{url}

\usepackage[breaklinks=true,colorlinks,bookmarks=false]{hyperref}

\title{\bf Blockout: Dynamic Model Selection for \\ Hierarchical Deep Networks}

\author{Calvin Murdock}
\affil{Machine Learning Department, Carnegie Mellon University, {\ttfamily cmurdock@cs.cmu.edu}}
\author{Zhen Li}
\author{Howard Zhou}
\author{Tom Duerig}
\affil{Google Research, {\ttfamily \{zhenli,howardzhou,tduerig\}@google.com}}
\date{} 

\begin{document}

\maketitle

\begin{abstract}

Most deep architectures for image classification--even those that are trained to classify a large number of diverse categories--learn shared image representations with a single model. Intuitively, however, categories that are more similar should share more information than those that are very different. While hierarchical deep networks address this problem by learning separate features for subsets of related categories, current implementations require simplified models using fixed architectures specified via heuristic clustering methods. Instead, we propose Blockout, a method for regularization and model selection that simultaneously learns both the model architecture and parameters. A generalization of Dropout, our approach gives a novel parametrization of hierarchical architectures that allows for structure learning via back-propagation. To demonstrate its utility, we evaluate Blockout on the CIFAR and ImageNet datasets, demonstrating improved classification accuracy, better regularization performance, faster training, and the clear emergence of hierarchical network structures.

\end{abstract}

\section{Introduction}

Multi-class classification is an important problem in visual understanding with applications ranging from image retrieval to robot navigation. Due to the vast space of variability, the seemingly simple task of identifying the subject of a photograph is extremely difficult. While once restricted to small label sets and constrained image domains, recent advances in deep neural networks have allowed image recognition to be applied to real-world collections of photographs. Effective image classification with thousands of labels and datasets with  millions of images are now commonplace. However, as classification tasks become more involved, larger networks with more capacity are required, emphasizing the importance of careful model selection. While much manual engineering effort has been dedicated to the task of designing deep architectures that are able to effectively generalize from available training data, model selection is typically performed using subjective heuristics by experienced practitioners. Furthermore, an appropriate architecture is closely tied to the dataset on which it is trained, so this work often must be repeated for each new application. Ideally, then, model selection should be performed automatically allowing the architecture to adapt to training data. As a step towards this goal, we propose an automated, end-to-end system for model selection within the class of hierarchical deep networks, which have demonstrated excellent performance on large-scale image classification tasks. 

\begin{figure}
\centering
\hspace*{\fill}
\hspace*{\fill}
\subfloat[]{
	\includegraphics[height=3cm]{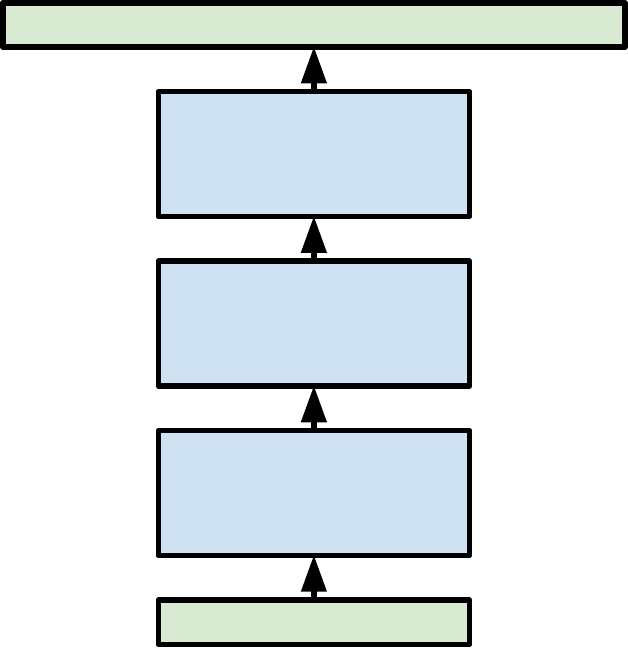}
}
\hspace*{\fill}
\subfloat[]{
	\includegraphics[height=3cm]{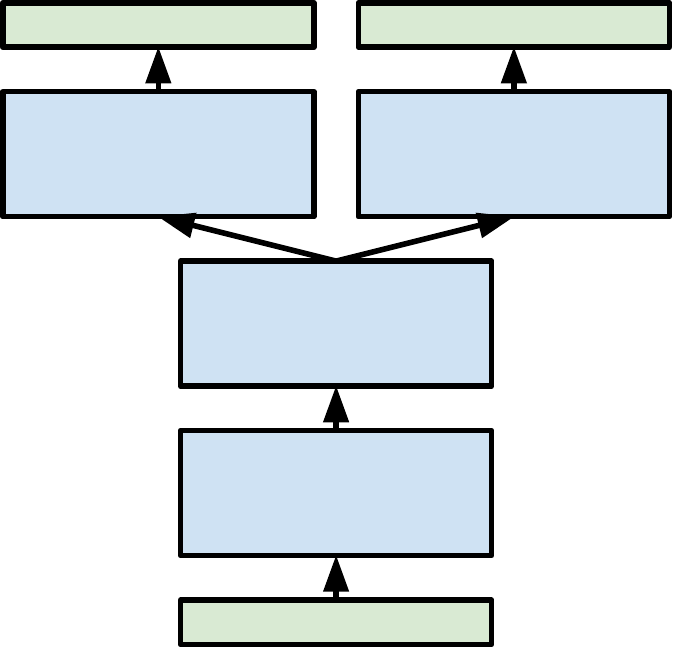}
}
\hspace*{\fill}
\subfloat[]{
	\includegraphics[height=3cm]{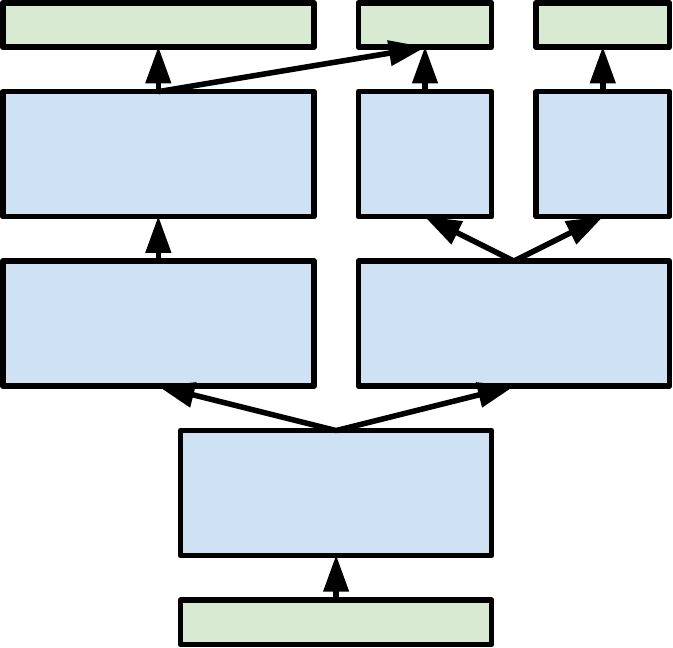}
}
\hspace*{\fill}
\hspace*{\fill}
\caption{ Example deep network architectures for multi-class classification that can be learned using Blockout. Input and output nodes are shown in green, groups of layers are shown in blue, and arrows indicate connections between them. (a) Traditional architectures make use of a combined model that computes a single feature vector for predicting a large number of diverse categories. (b) Hierarchical architectures instead partition the output categories into clusters and learn separate, high-level feature vectors for each of them. (c) Unlike previous approaches, Blockout allows for end-to-end learning of more complex hierarchical architectures. \vspace*{\fill} }
\label{fig:front}
\end{figure}

Deep neural networks are known to be organized such that specificity increases with depth~\cite{lee2009convolutional,yosinski2014transferable}. Lower layers tend to represent general, low-level image features like lines and edges while higher layers encode higher-level concepts like object parts or even objects themselves~\cite{coates2012emergence}. Most classification architectures make use of a single shared model with a flat logistic loss layer. Intuitively, however, categories that are more visually similar should share more information than those that are very different. 
For example, fine-grained features that are useful for differentiating between dog breeds likely differ from those useful for car models. 

One solution is to train independent fine-grained models for these subsets of related labels. This results in specialized features that are tuned to differentiating between subtle visual differences between similar categories. However, this is often infeasible due to limited training examples. On the other hand, a combined model for classifying many categories is able to use information common to all training images to learn shared, low-level representations. 

Ideally, then, model architectures should be hierarchical. Low-level representations should be shared while higher layers should be separated out and connected only to subsets of classes, allowing for efficient information sharing and reduced training data requirements. However, this raises an obvious question of model selection: which hierarchical architecture is best? Figure~\ref{fig:front} visualizes some potential candidates. Design choices include: the number and locations of branches, the allocation of nodes to each branch, and the clustering of classes in the final layer. 
Previous approaches to hierarchical deep networks (e.g. \cite{yan2014hd,warde2014self}) have simplified this question by fixing the base architecture and using heuristic clustering methods for separating the classes into groups based on their similarity. 
Unfortunately, the cumbersome disconnect between clustering and training means that model selection must rely on a different objective function that can only be correlated with the intended goal of maximizing prediction performance. 
While class similarity may provide an effective heuristic for this, it is not guaranteed to actually improve performance and ignores important factors such as heterogeneous classification difficulty
or the number of parameters that should be dedicated to each branch in the network. 

To achieve automatic model selection in hierarchical deep networks, we introduce Blockout, an approach for simultaneously learning both the model architecture and parameters. This allows for more complex hierarchical architectures specifically tuned to the data without requiring a separate procedure for model selection, which would likely be infeasible due to the vast search space of possible architectures. Inspired by Dropout~\cite{hinton2012improving}, Blockout can be viewed as a technique for stochastic regularization that adheres to hierarchically-structured model architectures. Importantly, its hyper-parameters (analogous to node Dropout probabilities) are represented such that they can be learned using simple back-propagation. 
Despite the additional parameters, the representational power of Blockout is exactly the same as a standard layer and can be parametrized as such during inference. Surprisingly, however, the resulting network is able to achieve improved performance, as demonstrated experimentally on standard image classification datasets. 

In summary, we make the following contributions: (1) a novel parametrization of hierarchical deep networks, (2) stochastic regularization analogous to Dropout that effectively averages over all models within this class of hierarchical architectures, (3) an approach for learning the regularization parameters allowing for architectures that dynamically adapt to the data throughout training, and (4) quantitative and qualitative analyses, including substantial performance gains over baseline models. 

\section{Related Work}

Despite the long history of deep neural networks in computer vision~\cite{lecun1998gradient}, the modern incarnation of ``deep learning'' is a relatively recent phenomenon that began with empirical success in the task of image recognition~\cite{krizhevsky2012imagenet}
on the ImageNet dataset~\cite{imagenet}. Since then, tactful architecture modifications have yielded a steady stream of further improvements~\cite{zeiler2014visualizing,szegedy2014inception}, even surpassing human performance~\cite{he2015delving}. 

In addition to general classification of arbitrary images, deep learning has also made a significant impact on fine-grained recognition within constrained domains~\cite{branson2014bird,krause2015fine,lin2015bilinear}. In these cases, deep neural networks are trained (often alongside additional annotations or segmentations of parts) to recognize subtle differences between similar categories, e.g. bird species. However, these methods are often limited by the availability of training data as they typically require expert annotations for ground truth labels.
Some approaches have alleviated this problem by pre-training on large collections of general images and then fine-tuning on smaller, domain-specific datasets~\cite{lin2015bilinear}. However, learning separate models for many different groups of categories would be inefficient.

Attempts have also been made to incorporate information from a known hierarchy to improve prediction performance without requiring architecture changes. For example, \cite{deng2014large} replaced the flat softmax classification layer with a probabilistic graphical model that respects given relationships between labels. Other methods for incorporating label structure are summarized in~\cite{tousch2012semantic}. However, they typically rely on fixed, manually-specified hierarchies, which could contain errors and result in biases that reduce performance. 

Hierarchical deep networks~\cite{warde2014self,yan2014hd} attempt to address these issues by learning multi-task models with shared lower layers and parallel, domain-specific higher layers for predicting different subsets of categories. While these methods address one component of model selection by learning clusters of output categories, other architectural hyper-parameters such as the location of branches and the relative allocation of nodes between them must still be specified prior to training.
Furthermore, these methods require separate steps for clustering and training while our approach automatically addresses these modeling choices in a single stage of end-to-end training.

The most common approach for model selection in deep learning is simply searching over the space of hyper-parameters~\cite{bergstra2012random}. Unfortunately, because training and inference in deep networks are computationally expensive, this is often impractical. While costs can sometimes be reduced (e.g. by taking advantage of the behavior of some network architectures with random weights~\cite{saxe2011random}), they still require training and evaluating a large number of models. Bayesian optimization approaches~\cite{snoek2012practical} attempt to perform this search more efficiently, but they  are still typically applied only to smaller models with few hyper-parameters. Alternatively,  \cite{arora2014provable} proposed a theoretically-justified approach to learning a deep network with a layer-wise strategy that automatically selects the appropriate number of nodes during training. However, it is unclear how it would perform on large-scale image classification benchmarks. 

A parallel but related task to model selection is regularization. A network with too much capacity (e.g. with too many parameters) can easily overfit without sufficient training data resulting in poor generalization performance. While the size of the model could be reduced, an easier and often more effective approach is to use regularization. Common methods include imposing constraints on the weights (e.g. through convolution or weight decay), rescaling or whitening internal representations for better conditioning~\cite{desjardins2015natural,ioffe2015batch}, or randomly perturbing activations for improved robustness and better generalizability~\cite{hinton2012improving,wan2013regularization,wager2013dropout}.

\section{Deep Neural Networks}

Deep neural networks are nonlinear functions $f:\mathbb{R}^{d}\rightarrow\mathbb{R}^{p}$ that take $d$-dimensional images as input and output $p$-dimensional predictions. They have been found to be very successful for image classification, most likely due to the complexity of the class of representable functions along with their ability to effectively and efficiently make use of very large sets of training data. 

Most deep neural networks are simply compositions of alternating linear and nonlinear functions.
More concretely, consider a deep network with $m$ layers. Each layer consists of a linear transformation $\boldsymbol{g}_j(\boldsymbol{x})=\mathbf{W}_{j}\boldsymbol{x}$ parametrized by $\mathbf{W}_{j}$ followed by a fixed nonlinear function $\boldsymbol{a}_j(\boldsymbol{x})$, e.g. a nonlinear activation or a pooling operator. Altogether, the full neural network can be represented as:
\begin{equation}
f=\boldsymbol{a}_{m}\circ\boldsymbol{g}_{m}\circ\boldsymbol{a}_{m-1}\circ\boldsymbol{g}_{m-1}\circ\cdots\circ\boldsymbol{a}_{1}\circ\boldsymbol{g}_{1}
\label{eq:func}
\end{equation}

Similarly, hierarchical deep networks can be expressed with a separate function $f$ for each subset of outputs where some intermediate representations $\boldsymbol{a}_j$ are shared, i.e. they can be used as the inputs to multiple layers.

The set of all model parameters $\mathbf{W}=\left\{\mathbf{W}_j\right\}$ can be learned from a dataset of $n$ training images $\boldsymbol{x}_i$ and corresponding ground-truth label vectors $\boldsymbol{y}_i$ using standard empirical risk minimization with a loss function $L$ (e.g. softmax) that measures the discrepancy between $\boldsymbol{y}_i$ and the network predictions $f(\boldsymbol{x}_i;\mathbf{W})$, as shown in Equation~\ref{eq:opt}:
\begin{equation}
\underset{\mathbf{W}}{\arg\min}\frac{1}{n}\sum_{i=1}^{n}L\big(\boldsymbol{y}_{i},\, f(\boldsymbol{x}_{i};\mathbf{W})\big)\quad\mbox{s.t. }\left\{\mathbf{W}_{j}\in\mathcal{S}_{j}\right\}
\label{eq:opt}
\end{equation}
Learning is typically accomplished through stochastic gradient descent, where the gradients of intermediate layers are computed using back-propagation.

Consistent with their name, deep neural networks typically consist of many layers that produce high-dimensional intermediate representations, resulting in an extremely large number of parameters to be learned. To prevent overfitting, regularization is typically employed through constraint sets $\mathcal{S}_j$ on the parameter matrices. 
The most common and effective form of regularization is convolution, which takes advantage of the local correlations of images and essentially restricts that the weight matrices contain shared parameters with a specific Toeplitz structure, resulting in far fewer free parameters to learn. Other examples of regularization include weight decay, which penalizes the norm of the weights, and Dropout, which has been shown (under certain assumptions) to indirectly impose a penalty function through stochastic perturbations of the internal network activations~\cite{wager2013dropout}. Blockout employs a similar form of stochastic regularization with the additional restriction that the parameter matrices be block-structured leading to hierarchical network architectures. 

\section{Hierarchical Network Parametrization} \label{sec:hierarchy}

\begin{figure}
\centering
\hspace*{\fill}
\hspace*{\fill}
\subfloat{
	\includegraphics[width=0.15\columnwidth]{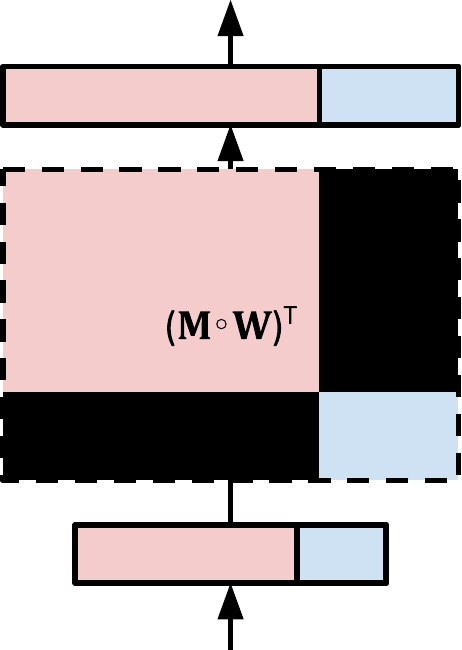}
}
\hspace*{\fill}
\subfloat{
	\includegraphics[width=0.15\columnwidth]{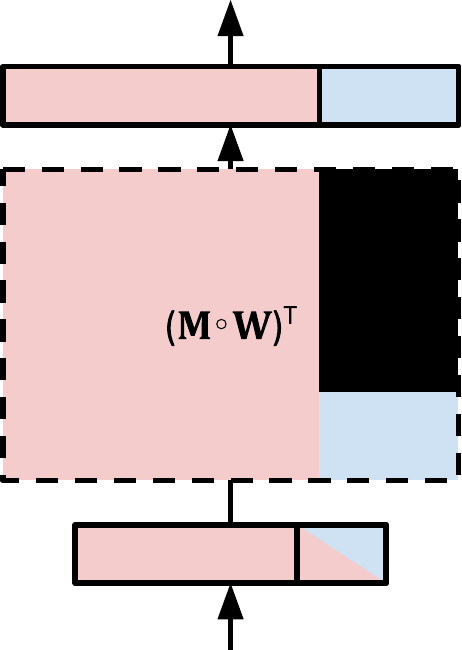}
}
\hspace*{\fill}
\subfloat{
	\includegraphics[width=0.15\columnwidth]{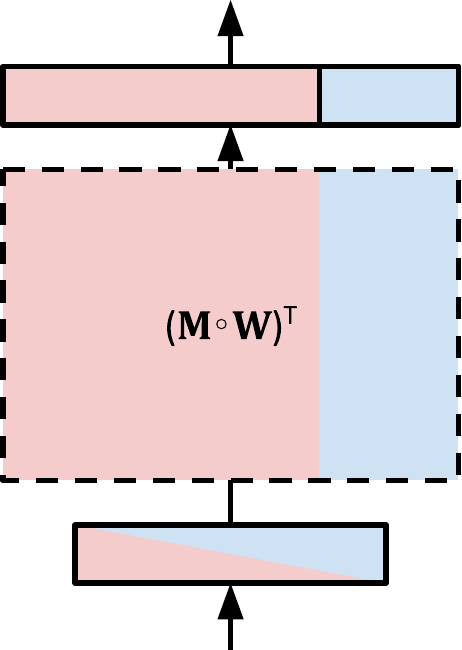}
}
\hspace*{\fill}
\hspace*{\fill}

\vspace{-1em}
\setcounter{subfigure}{0}
\hspace*{\fill}
\hspace*{\fill}
\subfloat[]{
	\includegraphics[width=0.15\columnwidth]{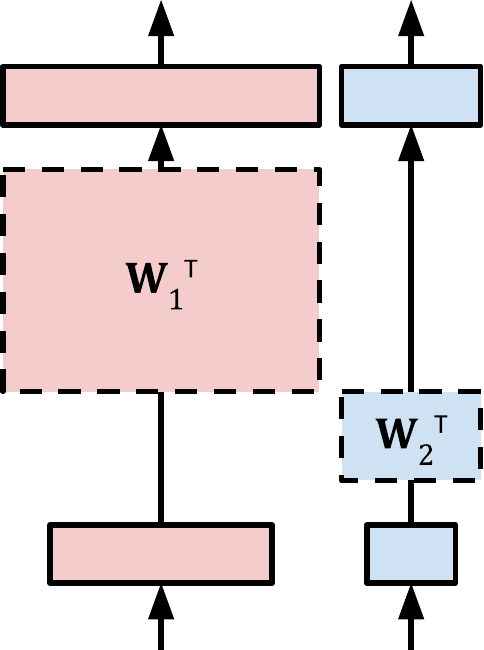}
}
\hspace*{\fill}
\subfloat[]{
	\includegraphics[width=0.15\columnwidth]{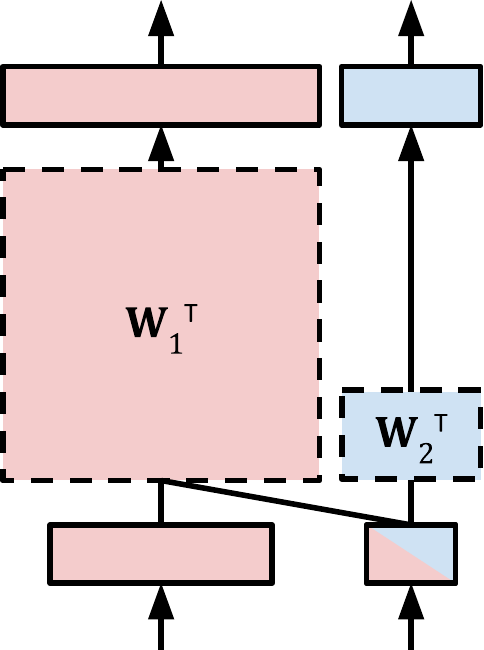}
}
\hspace*{\fill}
\subfloat[]{
	\includegraphics[width=0.15\columnwidth]{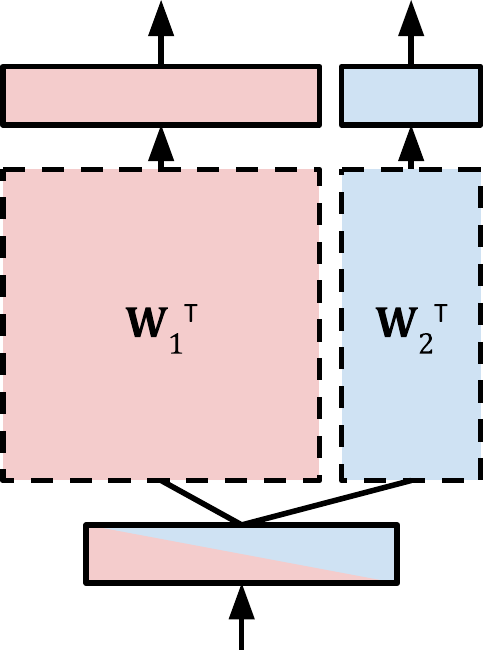}
}
\hspace*{\fill}
\hspace*{\fill}
\caption{ An illustration of the equivalence between single layers with block-structured parameter matrices (top) and parallel layers over subsets of nodes (bottom). Solid boxes indicate groups of nodes, dotted boxes represent the corresponding parameter matrices (where zero values are shown in black), and colors indicate cluster membership. (a) Independence between layers can be achieved when nodes only belong to a single cluster. When nodes belong to multiple clusters, hierarchical connections such as merging (b) and branching (c) can be achieved.  }
\label{fig:blockout}
\end{figure}

Blockout is based on the observation that parallel, independent layers can be equivalently expressed as a single combined layer with a block-structured weight matrix (up to a permutation of its rows and columns), as visualized in Figure~\ref{fig:blockout}. Thus, enforcing that the learned weight matrix have this type of structure during training automatically separates the input and output nodes into independent branches of a hierarchical architecture.

\begin{figure}
\centering
\hspace*{\fill}
\subfloat[Parallel]{
	\includegraphics[width=0.1\columnwidth]{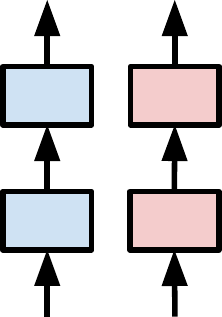}
}
\hspace*{\fill}
\subfloat[Dropout]{
	\includegraphics[width=0.1\columnwidth]{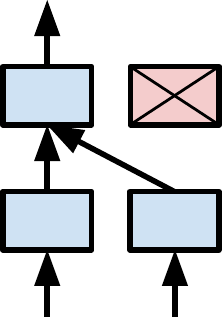}
}
\hspace*{\fill}
\subfloat[Dropout]{
	\includegraphics[width=0.1\columnwidth]{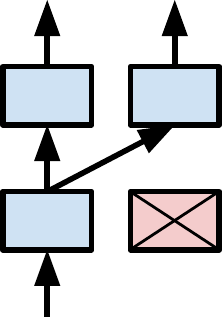}
}
\hspace*{\fill}
\subfloat[Branch]{
	\includegraphics[width=0.1\columnwidth]{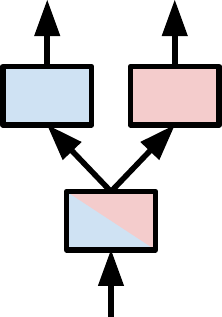}
}
\hspace*{\fill}
\subfloat[Merge]{
	\includegraphics[width=0.1\columnwidth]{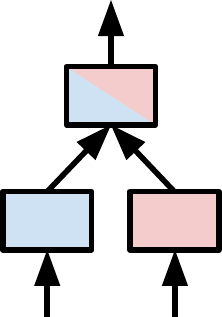}
}
\hspace*{\fill}
\caption{ A summary of the types of basic high-level architecture components that can be represented with Blockout. For each (a-e), a single layer is shown where groups nodes are shown as solid boxes, cluster assignments as colors, and connections within clusters as arrows. These connections allow for a rich space of potential model architectures. }
\label{fig:building_blocks}
\end{figure}

This can be parametrized by assigning each node to any number of $k$ clusters and masking out parameters if their corresponding input and output nodes do not belong to the same cluster, thus restricting the information that can be shared between nodes. Here, $k$ represents the maximum number of blocks in the parameter matrix or, equivalently, the maximum number of independent branches in the network. Though simple, this parametrization can encode a wide range of hierarchical structures, as shown in Figure~\ref{fig:building_blocks}.

More formally, we mask the parameter corresponding to $s^\mathrm{th}$ input node and the $t^\mathrm{th}$ output node as follows in Equation~\ref{eq:mask}, where $\widetilde{w}_{t,s}$ is the original, unconstrained parameter value and $\mathbb{I}(s\in C_{l})$ equals one if node $s$ belongs to cluster $l$ and zero otherwise:
\begin{equation}
\label{eq:mask}
w_{t,s}=\frac{1}{k}\sum_{l=1}^{k}\mathbb{I}(s\in C_{l})\mathbb{I}(t\in C_{l})\widetilde{w}_{s,t}
\end{equation}

This encodes the desired behavior that a parameter be nonzero only if its corresponding input and output nodes belong to the same class while restricting that the mask be between zero and one. Let $\mathbf{C}_{j}\in\left\{ 0,1\right\} ^{d_{j}\times k}$ be a binary indicator matrix containing these cluster membership assignments for each of the $d_j$ nodes in the output of the $j^\mathrm{th}$ layer. In other words, $\mathbf{C}_{j}(s,l)=\mathbb{I}(s\in C_{l})$. A full mask can then be constructed as $\frac{1}{k}\mathbf{C}_{j}\mathbf{C}_{j-1}^{\intercal}$ where the block-structured parameter matrix is the element-wise product of an unconstrained parameter matrix $\widetilde{\mathbf{W}}_{j}$ and this mask. This class of hierarchical architectures can be summarized by the constraint set in Equation~\ref{eq:constraint_set}, where $\odot$ indicates the element-wise Hadamard product.
\begin{equation}
\mathcal{S}_{j}=\left\{ \mathbf{W}_{j}:\mathbf{W}_{j}=\frac{1}{k}\widetilde{\mathbf{W}}_{j}\odot\mathbf{C}_{j}\mathbf{C}_{j-1}^{\intercal}\right\} 
\label{eq:constraint_set}
\end{equation}

These constraints act as a regularizer that enforces that the parameter matrices be block-structured with potentially many parameters set explicitly to zero. Ideally, we seek to learn the hierarchical structure during training, which is equivalent to learning the cluster membership assignments $\mathbf{C}_{j}$. However, because they are binary variables, learning them directly would be difficult. 
To address this problem, we instead take an approach akin to stochastic regularization approaches like Dropout: we treat cluster membership assignments as Bernoulli random variables and draw a different hierarchical architecture at each training iteration.

\section{Stochastic Regularization} \label{sec:stochastic_regularization}

Stochastic regularization techniques are simple but effective approaches for reducing overfitting in deep networks by injecting noise into the intermediate activations or parameters during training. Examples include Dropout~\cite{hinton2012improving}, which randomly sets activations to zero, and DropConnect~\cite{wan2013regularization}, which randomly sets parameter values to zero.
 
Dropout~\cite{hinton2012improving} works by setting node activations to zero with a certain probability at each training iteration. Inference is accomplished by replacing each activation with its expected value, which amounts to rescaling by the Dropout probability. This procedure approximates an ensemble of different models from the class of network architectures containing all possible subsets of nodes, where the Dropout probability determines the weight given to each architecture in this implicit model average. For example, with a high Dropout probability, models with fewer nodes are more likely to be selected during training. In general, Dropout results in improved generalization performance by preventing the coadaptation of features. 

Similarly, DropConnect~\cite{wan2013regularization} randomly sets parameter values to zero, which drops connections between nodes instead of the node activations themselves. During inference, a moment-matching procedure is used to better approximate an average over model architectures. Again, the success of this approach can be explained through its approximation of an ensemble within a much larger class of architectures: those that contain all possible combinations of connections between nodes in the network. 

Blockout can be seen as another example of stochastic regularization that approximates an ensemble of models from the class of hierarchical architectures introduced in Section~\ref{sec:hierarchy}. Structured noise is introduced by randomly selecting cluster assignments $\mathbf{C}_{j}$ corresponding to different hierarchical architectures at each iteration during training. We first consider the case of a single fixed probability $p$ that each node belongs to each of the clusters, but in Section~\ref{sec:backprop} we show how separate cluster probabilities can be learned for each node. 

During inference, we take an approach similar to Dropout and approximate an ensemble of hierarchical architectures using an implicit average with weights determined by the cluster probabilities. This again amounts to simply rescaling the parameter values by the expected value of the parameter mask: $p^2$.

\begin{figure}
\centering
\hspace*{\fill}
\subfloat[]{
	\includegraphics[height=3.3cm]{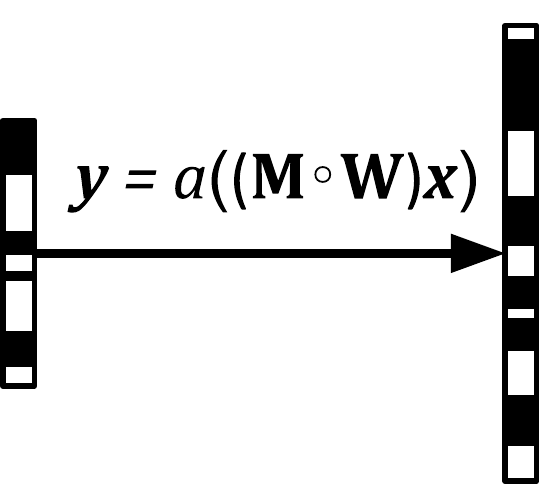}
}
\hspace*{\fill}
\subfloat[]{
	\includegraphics[height=3.3cm]{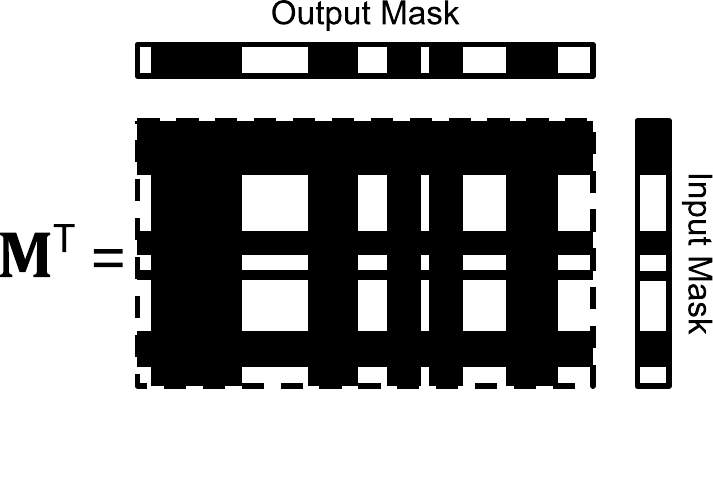}
}
\hspace*{\fill}
\caption{ A visualization of Dropout as applying a structured mask to the parameter matrix of a layer. (a) A single layer with Dropout applied to both its input ($\boldsymbol{x}$) and output ($\boldsymbol{y}$) nodes, which are represented as solid boxes with dropped activations shown in black. The computation is exactly the same as an ordinary layer, except with an element-wise mask applied to the weight matrix, as shown in (b). }
\label{fig:dropout}
\end{figure}

\begin{figure}
\centering
\hspace*{\fill}
\hspace*{\fill}
\subfloat[Dropout]{ 
	\includegraphics[width=0.2\columnwidth]{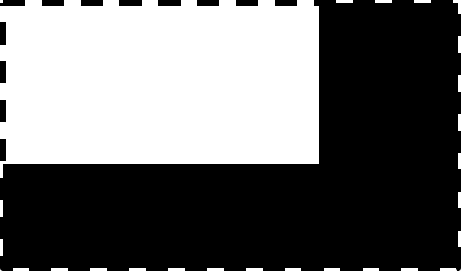}
}
\hspace*{\fill}
\subfloat[Blockout]{
	\includegraphics[width=0.2\columnwidth]{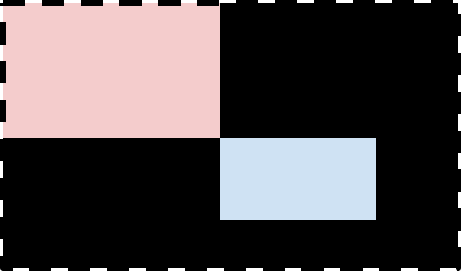}
}
\hspace*{\fill}
\subfloat[DropConnect]{
	\includegraphics[width=0.2\columnwidth]{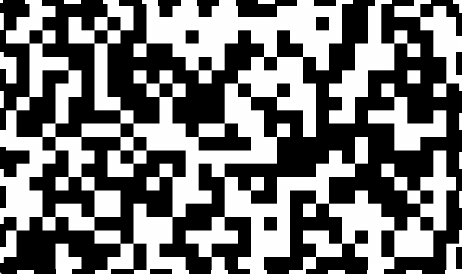}
}
\hspace*{\fill}
\hspace*{\fill}
\caption{ Example parameter masks that can be achieved with (a) Dropout, (b) Blockout, and (c) DropConnect. Note that Dropout and Blockout give block-structured, low-rank masks up to a permutation of the rows and columns while DropConnect is structureless, masking each parameter value independently.  }
\label{fig:masks}
\end{figure}

Also note that Dropout can be interpreted as implicitly applying a random mask $\mathbf{M}$ that sets parameters corresponding to the dropped inputs and outputs to zero. 
This is shown in Figure~\ref{fig:dropout}. 
If we reorder the input and output dimensions and permute the rows and columns of the weight matrix accordingly, the result is a single block of non-zero parameters, as shown in Figure~\ref{fig:masks}a. This is very similar to the block-structured masks that can be explicitly represented with Blockout, as shown in Figure~\ref{fig:masks}b. In fact, Dropout is equivalent to Blockout with $k=1$ where dropped nodes correspond to those that do not belong to the single cluster. In this case, the resulting mask is a rank-one matrix. Similarly, the explicit parameter masks in DropConnect (shown in Figure~\ref{fig:masks}c) are full-rank and can be equivalently represented by Blockout when the number of clusters is equal to the number of nodes in a layer. This allows each node to potentially belong to its own independent cluster resulting in a full-rank mask. 

The full intuition behind why stochastic regularization  approaches work and how best to select regularization hyper-parameters (e.g. Dropout probability) is lacking. On the other hand, Blockout gives a much clearer motivation: we assume that the output categories are hierarchically related, and so we approximate an ensemble only over hierarchical architectures. Furthermore, in Section~\ref{sec:backprop} we show how the cluster probabilities for each node can be learned from data allowing for the interpretation of Blockout as model selection within this class of architectures. 

\section{Learning Hierarchies via Back-Propagation} \label{sec:backprop}

The key difference between Blockout and other stochastic regularization techniques is that its hyper-parameters can be learned from data using simple back-propagation. To accomplish this, we replace the fixed, shared cluster probability $p$ with learnable parameters $\mathbf{P}_{j}\in\left[0,1\right]^{d_{j}\times k}$ whose elements represent the probability that each node belongs to each cluster. Essentially, they are relaxations of the binary cluster assignments $\mathbf{C}_{j}$ that can take on any value between zero and one and are implemented as real-valued variables followed by element-wise logistic activations. At each iteration of training, hard binary cluster assignments are drawn from Bernoulli distributions parametrized by these probabilities, i.e. $\mathbf{C}_{j}\sim\mbox{B}(1,\mathbf{P}_{j})$.

During training, the forward computations are performed using random masked weight matrices from the set in Equation~\ref{eq:constraint_set} for a different hierarchical architecture at each iteration. During inference, we again average over the cluster assignments to approximate an ensemble of hierarchical architectures. Since the cluster probabilities $\mathbf{P}_j$ are now different for each node, we must rescale each parameter accordingly. Specifically, the adjusted weight matrix used during inference is:
\begin{equation}
\mathbb{E}\left[\frac{1}{k}\widetilde{\mathbf{W}}_{j}\odot\mathbf{C}_{j}\mathbf{C}_{j-1}^{\intercal}\right]=\frac{1}{k}\widetilde{\mathbf{W}}_{j}\odot\mathbf{P}_{j}\mathbf{P}_{j-1}^{\intercal}
\end{equation}

Note that this leads to the same computation as that of the training forward pass except with $\mathbf{P}_{j}$ instead of $\mathbf{C}_{j}$. Thus, during inference, we simply skip the random cluster assignment step and use the soft clustering probabilities directly. 

\begin{figure}
\centering
\includegraphics[width=0.47\columnwidth]{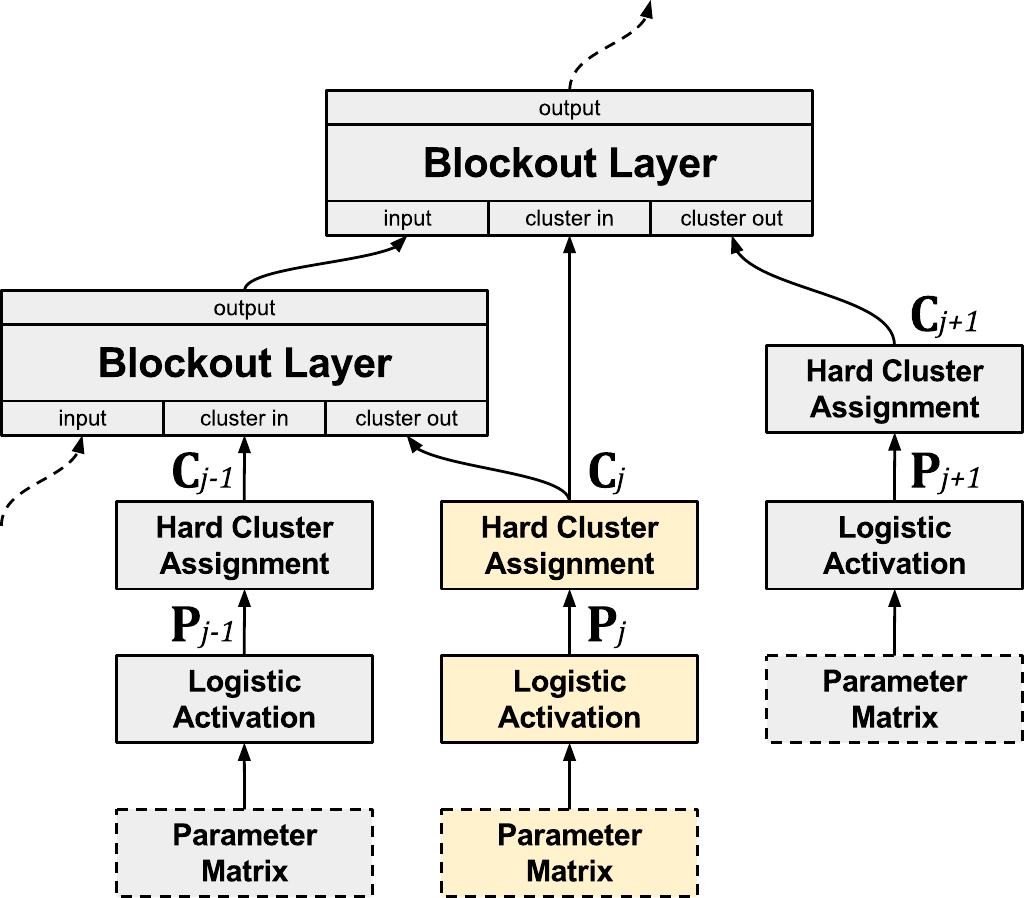}
\caption{ A block diagram showing two adjacent Blockout layers. 
Layers are shown as solid boxes, parameter matrices as dotted boxes, and connections as arrows. 
Note that the cluster membership parameters $\mathbf{C}_j$ applying to the  intermediate nodes (highlighted in orange) are used in the two adjacent layers and thus receive gradients from both. 
During inference, the hard cluster assignment layer is removed and the soft cluster probabilities are used directly. 
}
\label{fig:block_diagram}
\end{figure}

The masked parameter matrix is represented as a function of three variables: the unconstrained weight matrix $\mathbf{\widetilde{W}}_{j}$, the input cluster assignments $\mathbf{C}_{j-1}$, and the output cluster assignments $\mathbf{C}_{j}$. As such, gradients can be passed to all of them following the typical back-propagation algorithm. Specifically, updating $\mathbf{W}_{j}$ (e.g. using stochastic gradient descent) requires computation of the gradient of the loss function with respect to those parameters. Using the expression of a deep neural network as a composition of functions from Equation~\ref{eq:func}, this can be expressed using the chain rule as follows:
\begin{equation}
\frac{\partial L}{\partial\mathbf{W}_{j}}=\left(\frac{\partial L}{\partial\boldsymbol{a}_{m}}\frac{\partial\boldsymbol{a}_{m}}{\partial\boldsymbol{g}_{m}}\cdots\frac{\partial\boldsymbol{g}_{j+1}}{\partial\boldsymbol{a}_{j}}\frac{\partial\boldsymbol{a}_{j}}{\partial\boldsymbol{g}_{j}}\right)\frac{\partial\boldsymbol{g}_{j}}{\mathbf{W}_{j}}=\boldsymbol{\delta}_{j}\frac{\partial\boldsymbol{g}_{j}}{\mathbf{W}_{j}}
\end{equation}
where $\boldsymbol{\delta}$ is the product of all gradients from the loss function backwards down to the $j^\mathrm{th}$ layer. Using simple linear algebra, the gradients with respect to a layer's input (i.e. the previous layers activations $\boldsymbol{a}_{j-1}$) are:
\begin{equation}
\frac{\partial\boldsymbol{g}_{j}}{\partial\boldsymbol{a}_{j-1}}=\mathbf{W}_{j}=\frac{1}{k}\widetilde{\mathbf{W}}_{j}\odot\mathbf{C}_{j}\mathbf{C}_{j-1}^{\intercal}
\end{equation}
Similarly, the gradients with respect to all components of the weight matrix are computed as:
\begin{gather}
\label{eq:gradients}
\frac{\partial L}{\partial\mathbf{W}_{j}}=\boldsymbol{\delta}_{j}\boldsymbol{a}_{j-1}^{\intercal},\quad\frac{\partial L}{\partial\widetilde{\mathbf{W}}_{j}}=\frac{1}{k}\frac{\partial L}{\partial\mathbf{W}_{j}}\odot\mathbf{C}_{j}\mathbf{C}_{j-1}^{\intercal},
\\
\frac{\partial L}{\partial\mathbf{C}_{j}}=\frac{1}{k}\left[\widetilde{\mathbf{W}}_{j}\odot\frac{\partial L}{\partial\mathbf{W}_{j}}\right]^{\intercal}\mathbf{C}_{j-1}+\frac{1}{k}\left[\widetilde{\mathbf{W}}_{j+1}\odot\frac{\partial L}{\partial\mathbf{W}_{j+1}}\right]\mathbf{C}_{j+1} \notag
\end{gather}
Note that the cluster assignments for a set of nodes $\mathbf{C}_{j}$ are shared between the two adjacent Blockout layers and hence its gradient contains components from each. 
This can be seen pictorially in the block diagram in Figure~\ref{fig:block_diagram}.
This means that the effective parameters for a layer don't depend solely on information from the layer itself, 
acting as an additional form of regularization. 

Recall that our goal is to learn the cluster probabilities $\mathbf{P}_{j}$ that parametrize the cluster assignment random variables $\mathbf{C}_{j}$.  Thus, to update the cluster probabilities, we simply use the cluster assignment gradients after masking them so that the gradients of unselected clusters are zero:
\begin{equation}
\label{eq:gradients_2}
\frac{\partial L}{\partial\mathbf{P}_{j}}=\frac{\partial L}{\partial\mathbf{C}_{j}}\odot\mathbf{C}_{j}
\end{equation}
This is similar to the technique used when back-propagating gradients through a Dropout layer. 
Finally, to update the real-valued cluster parameters, these gradients are then back-propagated through the logistic activation layer. 
A block diagram of our implementation is shown in Figure~\ref{fig:block_diagram} and 
the full training process is summarized in Algorithm~\ref{alg:forward}. 

\begin{algorithm}[tb!]
\caption{ Blockout Training Iteration \label{alg:forward}}

\textbf{Input:} Mini-batch of training images $\left\{ \boldsymbol{x}_{i}\right\} _{i=1}^{B}$,
parameters from previous iteration $\widetilde{\mathbf{W}}_{j}^{\left(t-1\right)},\mathbf{P}_{j}^{\left(t-1\right)}$ 

\textbf{Output:} Updated parameters $\widetilde{\mathbf{W}}_{j}^{\left(t\right)},\mathbf{P}_{j}^{\left(t\right)}$

\vspace{0.2em}
\textbf{Forward Pass:} 
\begin{itemize}[noitemsep,nolistsep]
\item Draw cluster assignments: $\mathbf{C}_{j}\sim B(1,\mathbf{P}_{j}^{\left(t-1\right)})$
\item Mask parameters: $\mathbf{W}_{j}=\frac{1}{k}\widetilde{\mathbf{W}}_{j}^{\left(t-1\right)}\odot\mathbf{C}_{j}\mathbf{C}_{j-1}^{\intercal}$
\item Compute predictions: $\hat{\boldsymbol{y}}_{i}=f(\boldsymbol{x}_{i};\mathbf{W}_{j})$
\item Evaluate empirical risk: $\frac{1}{B}\sum_{i=1}^{B}L\big(\boldsymbol{y}_{i},\,\hat{\boldsymbol{y}}_{i}\big)$
\end{itemize}

\textbf{Backward Pass:}
\begin{itemize}[noitemsep,nolistsep]
\item Compute gradients according to Equations~\ref{eq:gradients} and~\ref{eq:gradients_2}. 
\item Update parameters $\widetilde{\mathbf{W}}_{j}^{\left(t\right)},\mathbf{P}_{j}^{\left(t\right)}$
accordingly.  
\end{itemize}
\end{algorithm}

Modifying Equation~\ref{eq:opt}, our final optimization problem can thus be written as follows:
\begin{equation}
\begin{gathered}
\underset{\widetilde{\mathbf{W}},\mathbf{P}}{\arg\min}\frac{1}{n}\sum_{i=1}^{n}\mathbb{E}_{\mathbf{C}\sim B(1,\mathbf{P})}L\big(\boldsymbol{y}_{i},\, f(\boldsymbol{x}_{i};\mathbf{W})\big)\\[-8pt]
\mbox{s.t. }\mathbf{W}_{j}=\frac{1}{k}\widetilde{\mathbf{W}}_{j}\odot\mathbf{C}_{j}\mathbf{C}_{j-1}^{\intercal}
\end{gathered}
\end{equation}
The expectation over all cluster assignments is approximated using stochastic regularization, which randomly samples a different hierarchical architecture at each iteration. The effective weights of each architecture in the implicit ensemble are determined by the cluster probabilities $\mathbf{P}_{j}$, which can easily be learned as described using back-propagation. 
These cluster probabilities are initialized to 0.5 for maximum uncertainty. 

Throughout training, there are a number of possible outcomes: (1) The probabilities could diverge, some towards one and others towards zero. This would result in a fixed clustering of nodes, giving high confidence to a particular learned hierarchical structure. (2) Alternatively, the gradients could be uninformative, averaging to zero and leading to unchanged probabilities. This could indicate that hierarchical architectures are helpful for regularization, but the particular grouping of nodes is arbitrary. (3) The probabilities could also all increase, possibly demonstrating that hierarchical architectures are not beneficial and better performance could be achieved with fully-connected layers.

\section{Experimental Results}

To evaluate our approach, we apply Blockout to the standard image classification datasets CIFAR~\cite{cifar} and ImageNet~\cite{imagenet}. As baselines, we use variations of the Inception architecture~\cite{szegedy2014inception}. Specifically, for ImageNet we use the same model described in~\cite{ioffe2015batch}, and for CIFAR we use a compacted version of this model with fewer layers and parameters. We also follow the same training details described in~\cite{ioffe2015batch} with standard data augmentation. These models have been hand-engineered to achieve very good, near state-of-the-art performance by themselves. Thus, our intention is 
to show how the addition of Blockout layers can easily improve performance without involved hyper-parameter tuning. Furthermore, we show that Blockout does indeed learn hierarchical network structures resulting in higher prediction accuracy and faster training.

\begin{figure}
\centering
\hspace*{\fill}
\hspace*{\fill}
\subfloat[]{
	\includegraphics[height=5.7cm]{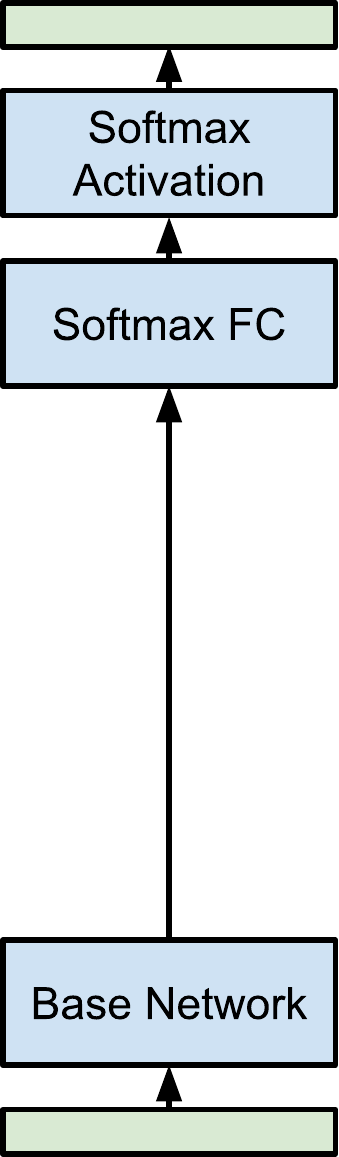}
}
\hspace*{\fill}
\subfloat[]{
	\includegraphics[height=5.7cm]{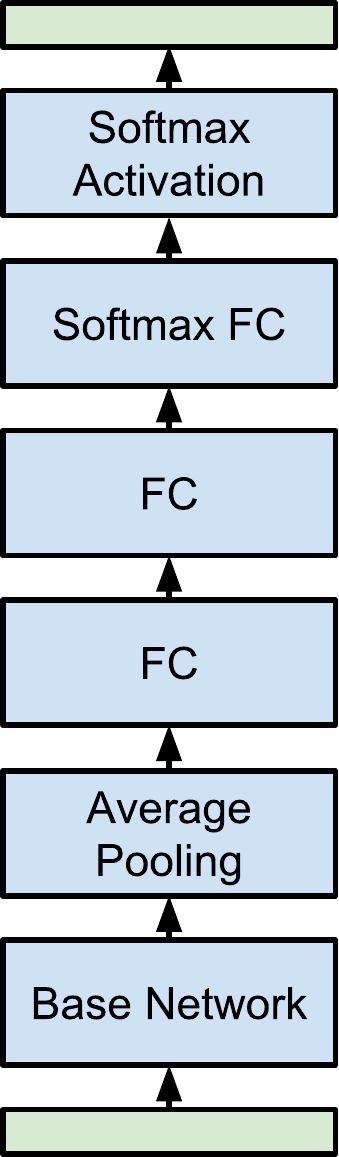}
}
\hspace*{\fill}
\subfloat[]{
	\includegraphics[height=5.7cm]{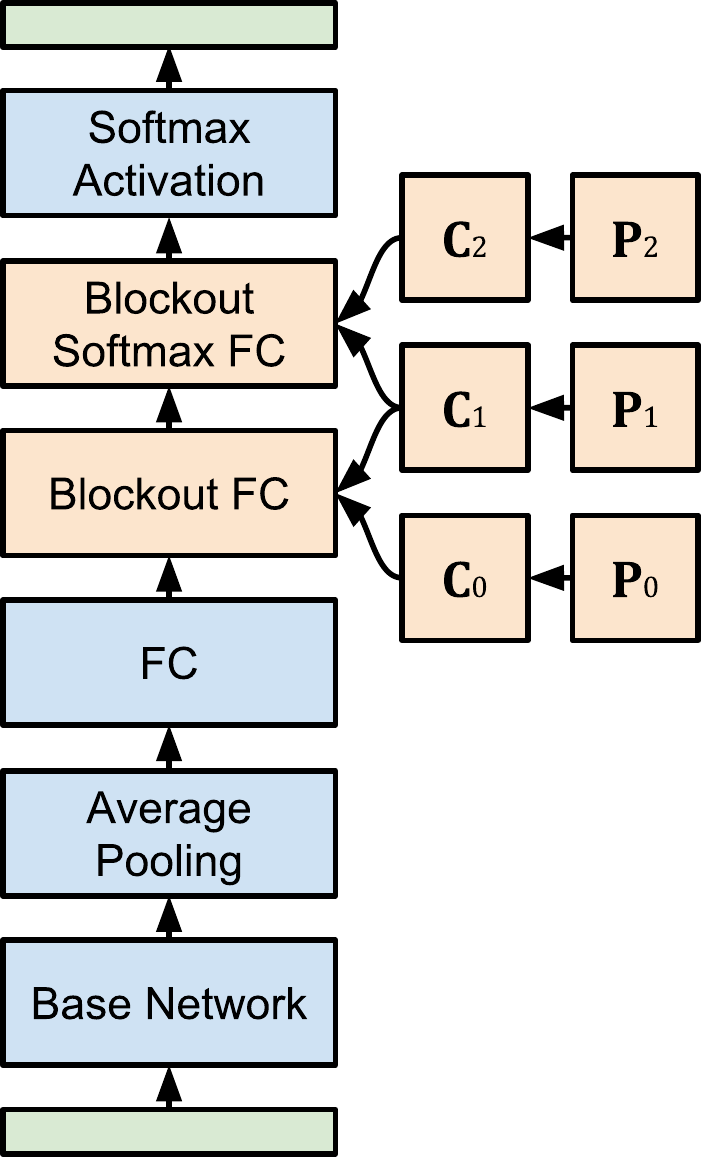}
}
\hspace*{\fill}
\hspace*{\fill}
\caption{ Block diagrams of the models compared in our experiments. (a) As baselines, we use variants of the Inception convolutional neural network architecture~\cite{szegedy2014inception}. (b) For comparison, we add an average pooling layer to reduce the bottleneck size followed by two fully-connected layers (potentially with Dropout) before the softmax classifier. (c) Our model replaces the last two FC layers with Blockout layers of the same size. }
\label{fig:architectures}
\end{figure}

Inception architectures are composed of multiple layers of parallel convolutional operations directly followed by softmax classification. However, it has been shown that fully-connected layers before classification act as a form of orderless pooling~\cite{lin2015bilinear} and have been used extensively~\cite{krizhevsky2012imagenet,zeiler2014visualizing,simonyan2014very} demonstrating improved model capacity leading to better performance. Thus, we add two fully-connected layers after the convolutional layers of our base architectures. Because our baselines already have high network capacity, doing this naively can lead to extreme overfitting and reduced performance, even with standard regularization techniques such as Dropout. However, using Blockout prevents this overfitting and leads to substantial performance improvements with a wide range of hyper-parameter choices. The architectures compared in our experiments are shown in Figure~\ref{fig:architectures}.

To demonstrate the effectiveness of the different components of Blockout, we compare three variations of our proposed model. The first, indicated by (soft, learned) in the following experiments, omits random cluster selection by skipping the Bernoulli sampling step (i.e. the hard cluster assignment layers in Figure~\ref{fig:block_diagram} are removed during both training and inference instead of just during inference.) 
This effectively removes the stochastic regularization effect of Blockout, instead using the relaxed soft clustering assignments directly by setting $\mathbf{C}_j=\mathbf{P}_j$. Without explicit zero-valued parameters, the same number of effective parameters are learned as with ordinary fully-connected layers, which could still potentially lead to over-fitting. However, the additional regularization provided by the shared cluster parameters often mitigates this, still resulting in improved performance. The second (hard, fixed) uses randomized hard cluster assignment during training, but uses fixed cluster probabilities of 0.5 instead of back-propagating gradients as described in Section~\ref{sec:backprop}. This shows the effects of stochastic regularization within the class of hierarchical architectures. Finally, the third (hard, learned) is our full proposed model.

\subsection{CIFAR-100}

\begin{figure}
\centering
\hspace*{\fill}
\subfloat[Training Cost]{
	\includegraphics[width=0.33\columnwidth]{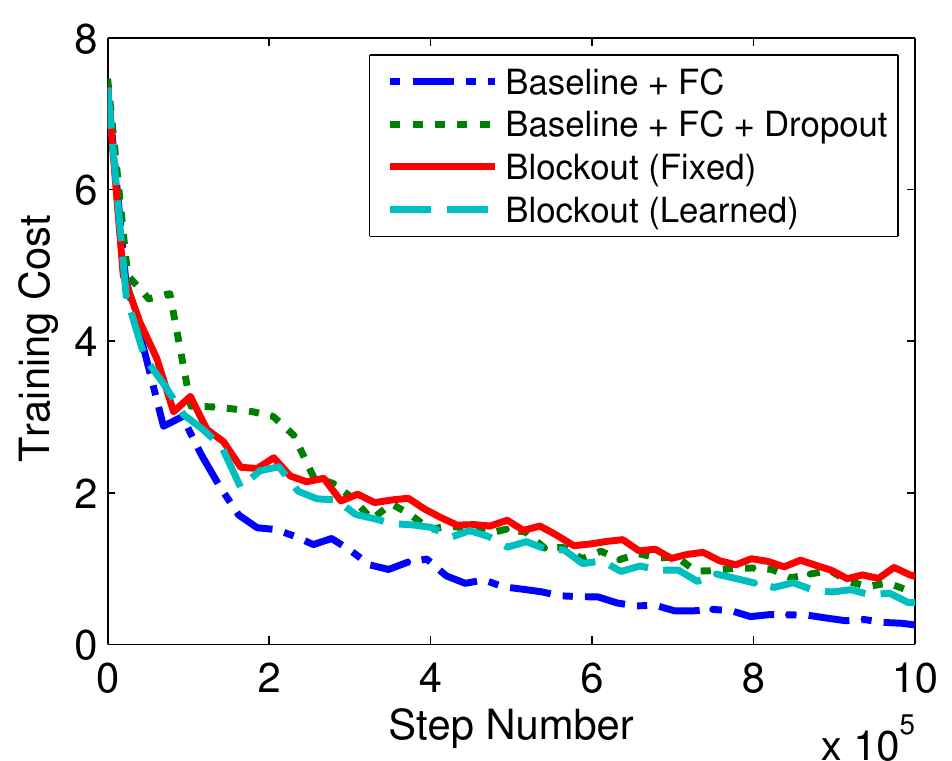}
}
\hspace*{\fill}
\hspace*{\fill}
\subfloat[Training Accuracy]{
	\includegraphics[width=0.33\columnwidth]{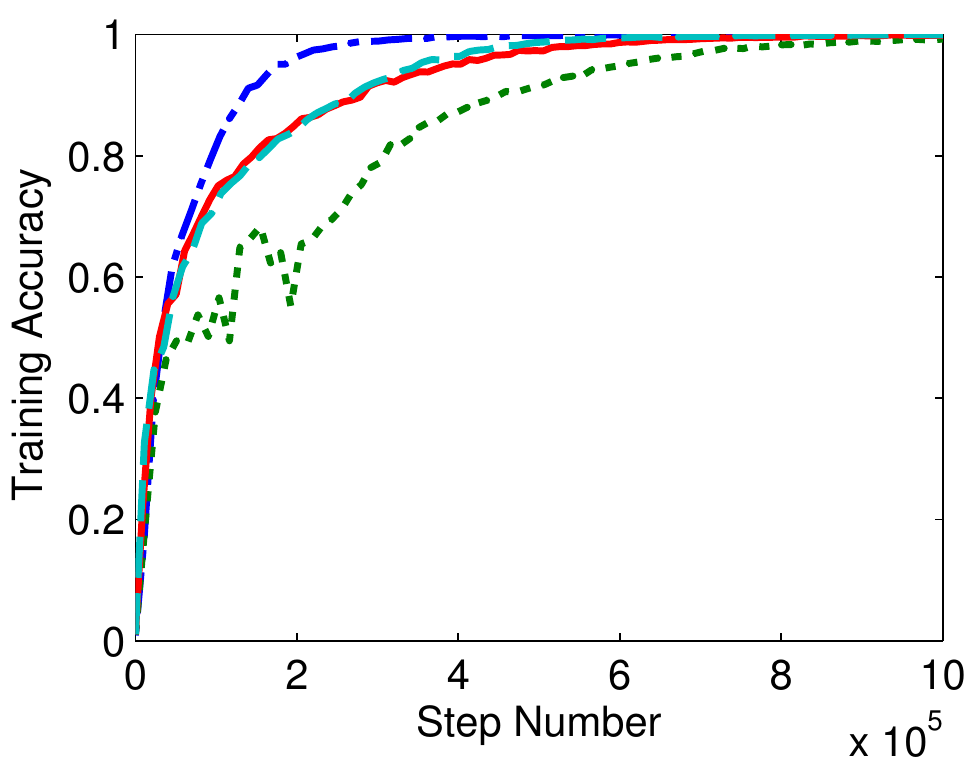}
}
\hspace*{\fill}
\subfloat[Testing Accuracy]{
	\includegraphics[width=0.33\columnwidth]{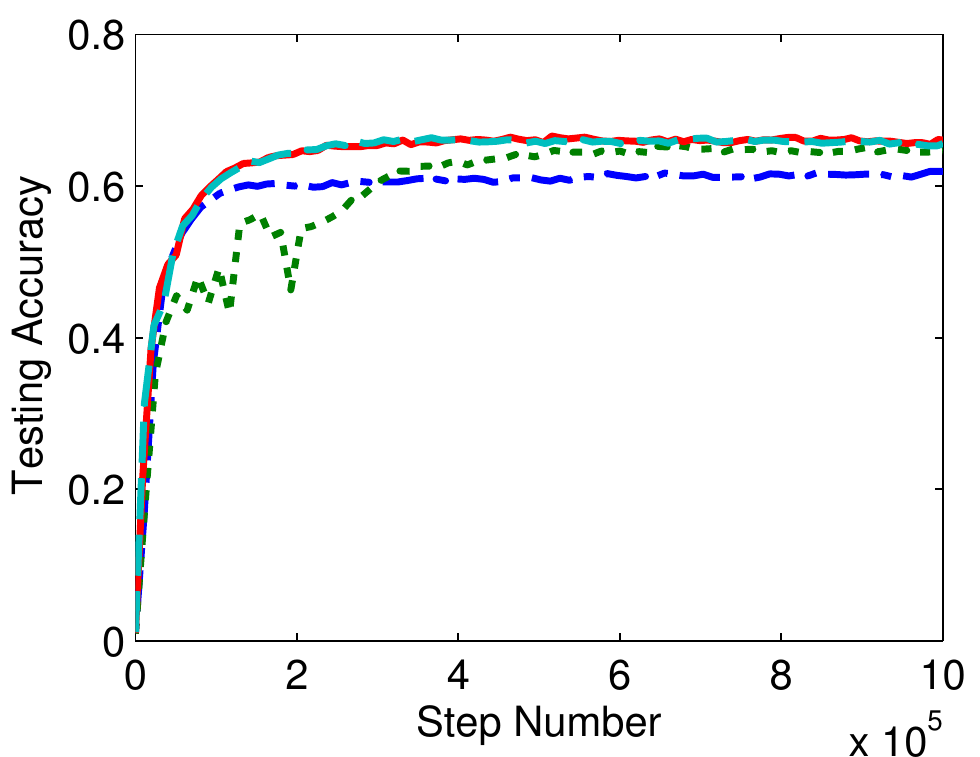}
}
\hspace*{\fill}
\hspace*{\fill}
\caption{ The convergence of our models in comparison to the baselines on the CIFAR-100 dataset, showing the training cost (a) and the accuracy on the (b) training and (c) testing sets throughout training. Note that Blockout converges in about half the time as Dropout while still achieving a higher final accuracy. }
\label{fig:cifar_convergence}
\end{figure}

CIFAR-100 is a challenging dataset comprising of 60k 32x32 color images (500k for training and 100k for testing) equally divided into 100 classes~\cite{cifar}.
Table~\ref{tab:cifar100_acc} shows the performance of our model with 6 clusters and 512 nodes in each fully-connected layer. We compare against the baseline Inception model, the baseline with fully-connected layers, and the baseline with fully-connected layers followed by 30\% Dropout. Figure~\ref{fig:cifar_convergence} shows the cost and accuracy of these models throughout training, demonstrating faster convergence in comparison to Dropout. Table~\ref{tab:cifar100_acc} also compares our full Blockout model (hard, learned) with a variety of hyper-parameter selections, including the number of hidden nodes in each fully-connected layer and the number of clusters. 

The best performance achieved by our method gave an accuracy of 66.71\% with 6 clusters and 2048 nodes, showing a significant improvement over the baseline accuracy. Also note that, while other stochastic regularization methods like Dropout can still overfit if there are too many parameters or the Dropout probability is not set correctly, Blockout seems to adapt so that adding more nodes never reduces accuracy. Despite its minimal engineering effort, the results are comparable to state-of-the-art methods (e.g. 68.8\% with~\cite{graham2014fractional}, 67.76\% with~\cite{srivastava2015training}, 66.29\% with~\cite{springenberg2014striving}, etc.)

\begin{table}
\centering
\caption{ {\bfseries Cifar-100 Test Accuracy.} Left: Comparison with baseline methods. Right: Variable clusters with 512 nodes (top) and variable nodes with 6 clusters (bottom).
\label{tab:cifar100_acc}
}
\begin{minipage}{0.3\columnwidth}
\centering
\begin{tabular}{|l|c|}
\hline 
\textbf{Method} & \textbf{Acc. (\%)}\tabularnewline
\hline 
\hline 
Baseline & 61.56\tabularnewline
\hline 
Baseline + FC & 62.66\tabularnewline
\hline 
Baseline + FC + Dropout & 64.32\tabularnewline
\hline 
\hline 
Blockout (soft, learned) & 63.57\tabularnewline
\hline 
Blockout (hard, fixed) & 65.62\tabularnewline
\hline 
Blockout (hard, learned) & \textbf{65.66}\tabularnewline
\hline 
\end{tabular}
\end{minipage} \hspace*{1em}
\begin{minipage}{0.3\columnwidth}
\centering
\begin{tabular}{|c|c|}
\hline 
\textbf{Clusters} & \textbf{Acc. (\%)}\tabularnewline
\hline 
\hline 
2 & 64.54\tabularnewline
\hline 
4 & \textbf{65.93}\tabularnewline
\hline 
6 & 65.66\tabularnewline
\hline 
\multicolumn{1}{c}{} & \multicolumn{1}{c}{}\tabularnewline
\hline 
\textbf{Nodes} & \textbf{Acc. (\%)}\tabularnewline
\hline 
\hline 
512 & 65.66\tabularnewline
\hline 
1024 & 66.69\tabularnewline
\hline 
2048 & \textbf{66.71}\tabularnewline
\hline 
\end{tabular}
\end{minipage}
\end{table}

\subsection{ImageNet}

ImageNet is the standard dataset for large-scale image classification~\cite{imagenet}. We use the version of the dataset from the Imagenet Large Scale Visual Recognition Challenge (ILSVRC 2012), which has 1000 object categories, 1.2 million training images, and 50k validation images. Table~\ref{tab:imagenet_acc} shows the top-1 prediction performance of our model with 6 clusters and 4096 nodes in each fully-connected layer. 
We compare to the baseline, the baseline with fully-connected layers, and the baseline with fully-connected layers followed by 50\% Dropout. 
Because the baseline model was already carefully tuned to maximize performance on ImageNet, adding fully-connected layers resulted in significant overfitting that could not be overcome with Dropout. However, Blockout was able to effectively remove these effects giving an improved final maximum performance of 74.95\%.

\begin{table}
\centering
\caption{ {\bfseries ImageNet Evaluation Accuracy.} Left: Comparison with baseline methods. Right: Variable clusters with 4096 nodes (top) and variable nodes with 6 clusters (bottom). \label{tab:imagenet_acc}
}
\begin{minipage}{0.3\columnwidth}
\centering
\begin{tabular}{|l|c|}
\hline 
\textbf{Method} & \textbf{Acc. (\%)}\tabularnewline
\hline 
\hline 
Baseline & 73.43\footnotemark{}\tabularnewline
\hline 
Baseline + FC & 68.06\tabularnewline
\hline 
Baseline + FC + Dropout & 73.88\tabularnewline
\hline 
\hline 
Blockout (soft, learned) & 72.43\tabularnewline
\hline 
Blockout (hard, fixed) & 74.44\tabularnewline
\hline 
Blockout (hard, learned) & \textbf{74.83}\tabularnewline
\hline 
\end{tabular}
\end{minipage} \hspace*{1em}
\begin{minipage}{0.3\columnwidth}
\centering
\begin{tabular}{|c|c|}
\hline 
\textbf{Clusters} & \textbf{Acc. (\%)}\tabularnewline
\hline 
\hline 
2 & 73.78\tabularnewline
\hline 
6 & \textbf{74.83}\tabularnewline
\hline 
15 & 74.19\tabularnewline
\hline 
\multicolumn{1}{c}{} & \multicolumn{1}{c}{}\tabularnewline
\hline 
\textbf{Nodes} & \textbf{Acc. (\%)}\tabularnewline
\hline 
\hline 
1024 & 74.16\tabularnewline
\hline 
2048 & 74.47\tabularnewline
\hline 
4096 & 74.83\tabularnewline
\hline 
8192 & \textbf{74.95}\tabularnewline
\hline 
\end{tabular}
\end{minipage}
\end{table}

\begin{table}
\centering
\caption{ {\bfseries Cifar-10 Test Accuracy.} \label{tab:cifar10_acc}
}
\begin{tabular}{|l|c|}
\hline 
\textbf{Method} & \textbf{Acc. (\%)}\tabularnewline
\hline 
\hline 
Baseline & 93.65\tabularnewline
\hline 
Baseline + FC & 93.39\tabularnewline
\hline 
Baseline + FC + Dropout & 93.52\tabularnewline
\hline 
\hline 
Blockout (soft, learned) & \textbf{94.41}\tabularnewline
\hline 
Blockout (hard, fixed) & 93.67\tabularnewline
\hline 
Blockout (hard, learned) & 93.55\tabularnewline
\hline 
\end{tabular}
\end{table}

\begin{figure}[t!]
\centering
\begin{minipage}{.87\columnwidth}
\centering
\hspace*{\fill}{\bf \quad $\mathbf{P}_0$}\hspace*{\fill}
\hspace*{\fill}{\bf \quad $\mathbf{P}_1$}\hspace*{\fill}
\hspace*{\fill}{\bf \quad $\mathbf{P}_2$}\hspace*{\fill}
\vspace{-1em}

\subfloat[Soft Clustering, 4096 Nodes]{
	\includegraphics[width=0.32\columnwidth]{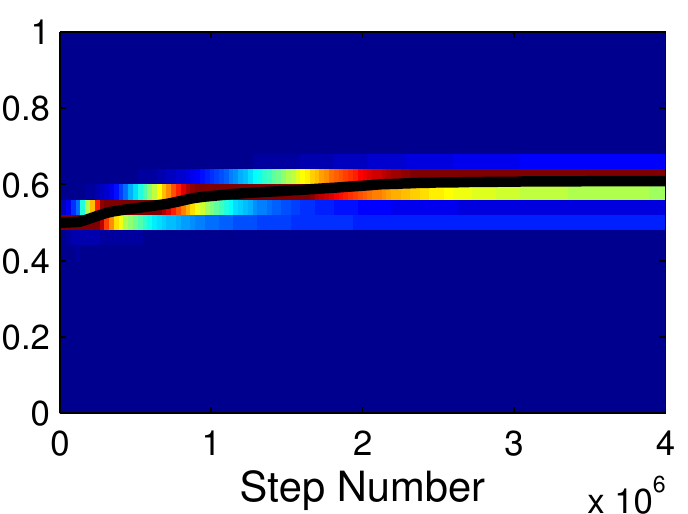}
	\includegraphics[width=0.32\columnwidth]{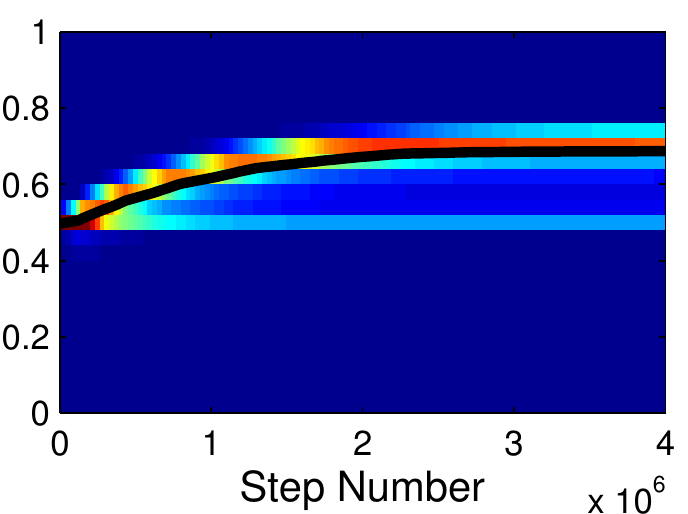}
	\includegraphics[width=0.32\columnwidth]{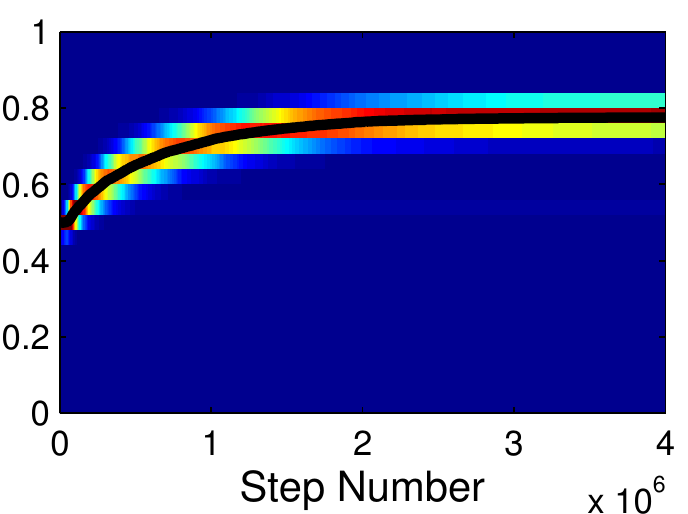}
}
\vspace{-1em}
\subfloat[Hard Clustering, 4096 Nodes]{
	\includegraphics[width=0.32\columnwidth]{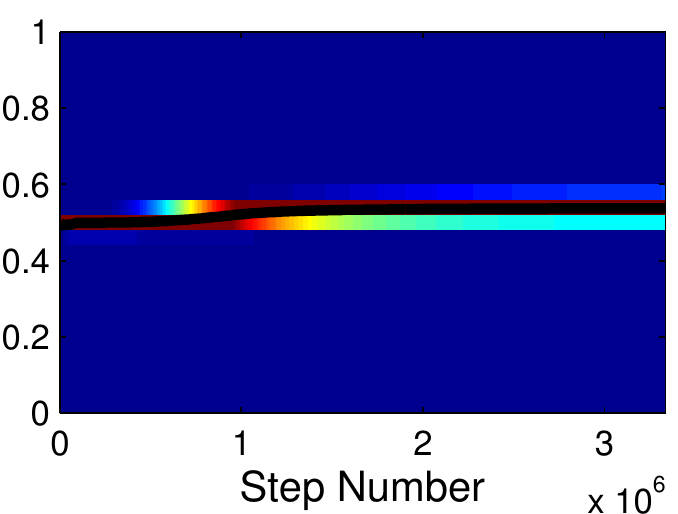}
	\includegraphics[width=0.32\columnwidth]{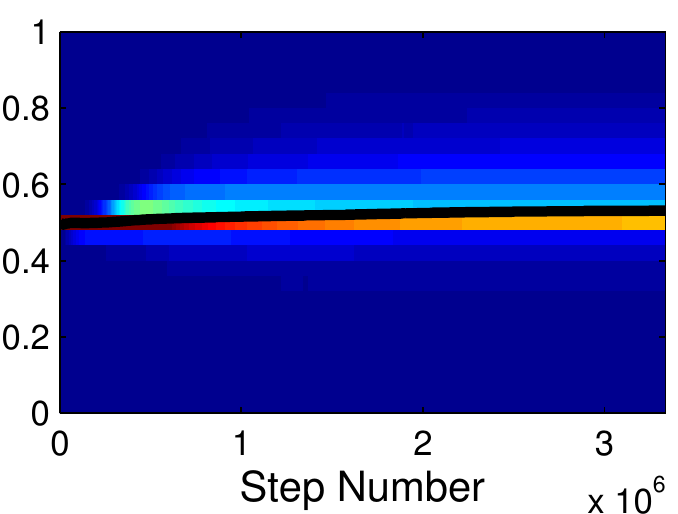}
	\includegraphics[width=0.32\columnwidth]{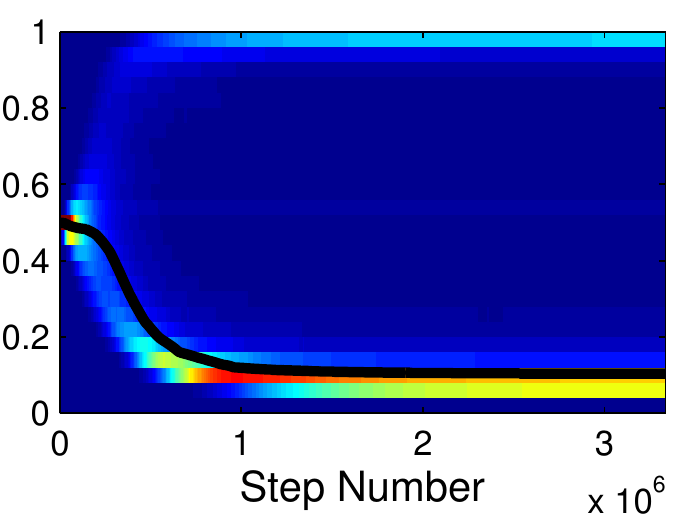}
}
\vspace{-1em}
\subfloat[Hard Clustering, 2048 Nodes]{
	\includegraphics[width=0.32\columnwidth]{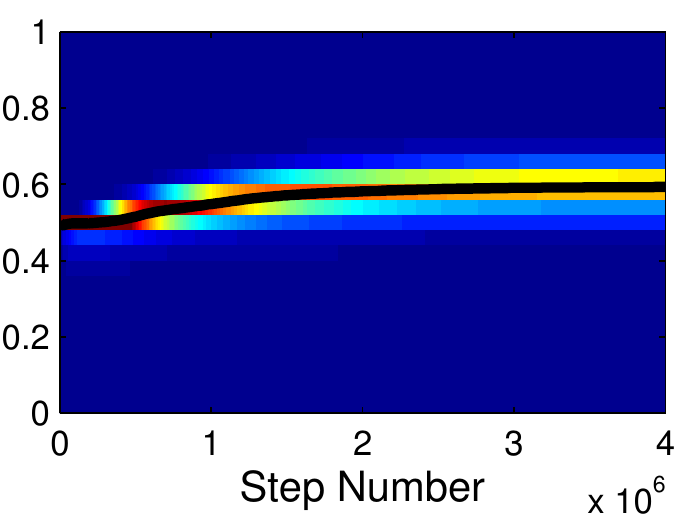}
	\includegraphics[width=0.32\columnwidth]{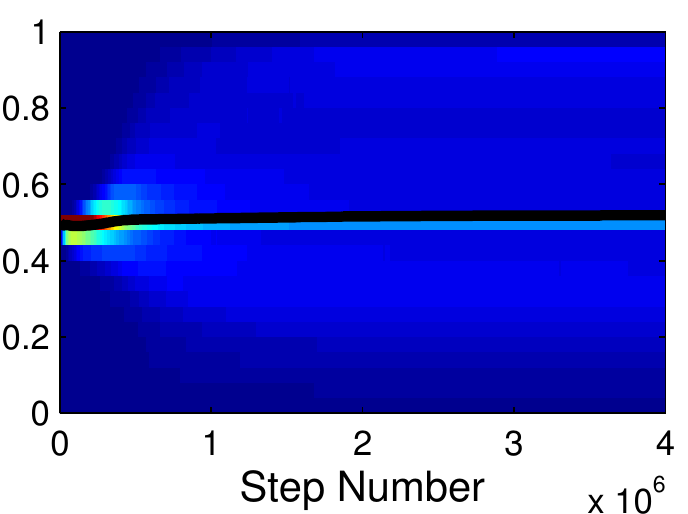}
	\includegraphics[width=0.32\columnwidth]{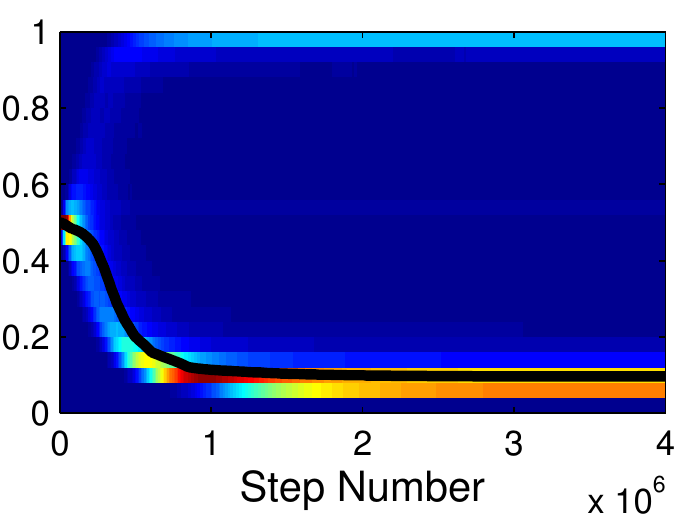}
}
\vspace{-1em}
\subfloat[Hard Clustering, 1024 Nodes]{
	\includegraphics[width=0.32\columnwidth]{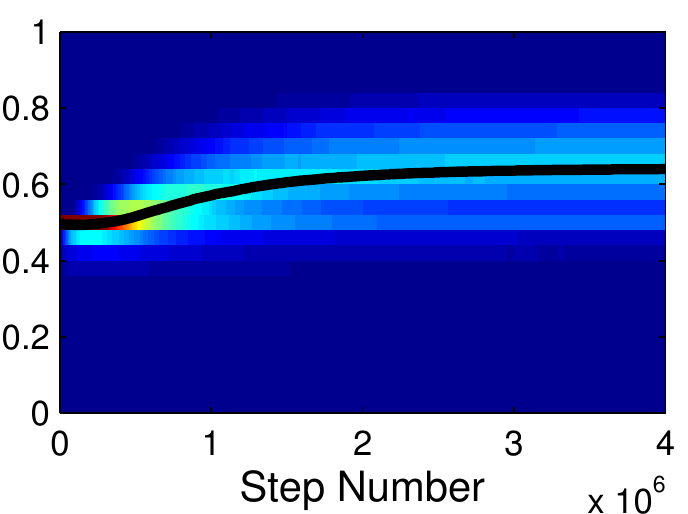}
	\includegraphics[width=0.32\columnwidth]{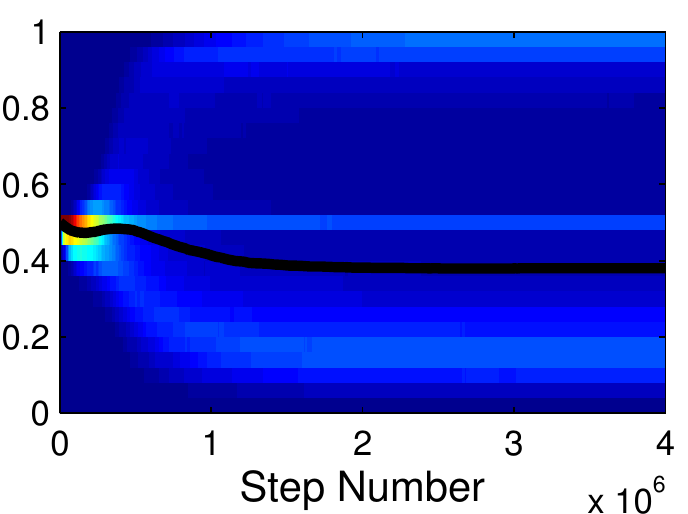}
	\includegraphics[width=0.32\columnwidth]{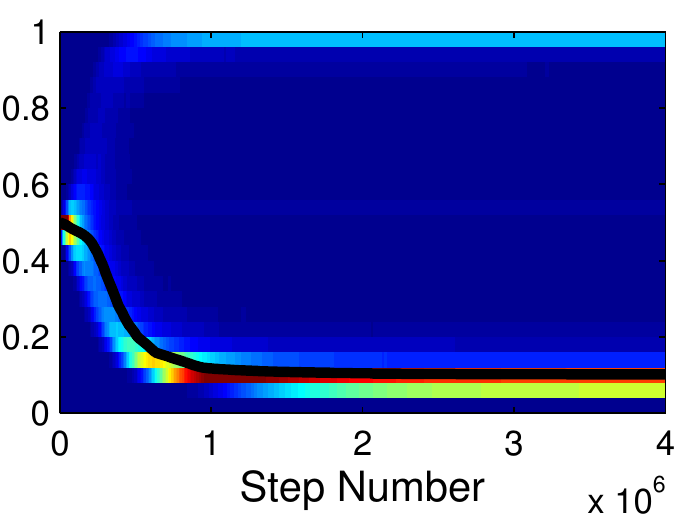}
}
\end{minipage}
\caption{ A visualization of the distributions of each layer's cluster probabilities $\mathbf{P}_j$ throughout training on the ImageNet dataset. The iteration number varies along the x-axis with probability along the y-axis. Warmer colors indicate a higher density of cluster probabilities at a given iteration while the black line shows their median. With hard clustering, there is a clear separation towards higher confidence cluster assignments, especially in later layers. \vspace{0.5em} }
 \label{fig:imagenet_histogram}
\end{figure}

Figure~\ref{fig:imagenet_histogram} shows the distribution of the learned cluster probabilities throughout training. Without random cluster selection, soft clustering causes all probabilities to increase towards one, which could indicate overfitting to the training data. On the other hand, stochastic regularization with hard clustering results in diverging probabilities giving higher confidence cluster membership assignments. This effect is also more prevalent in higher layers, agreeing with our intuition that more information should be shared between clusters in lower layers. 

\begin{figure}[t!]
\centering
\begin{minipage}{0.8\columnwidth}
\centering
\hspace*{\fill}{\bf \quad $\mathbf{P}_0$}\hspace*{\fill}
\hspace*{\fill}{\bf \quad $\mathbf{P}_1$}\hspace*{\fill}
\hspace*{\fill}{\bf \quad $\mathbf{P}_2$}\hspace*{\fill}

\vspace{-1em}
\subfloat[Hard Clustering, 1024 Nodes]{
\includegraphics[width=0.31\columnwidth]{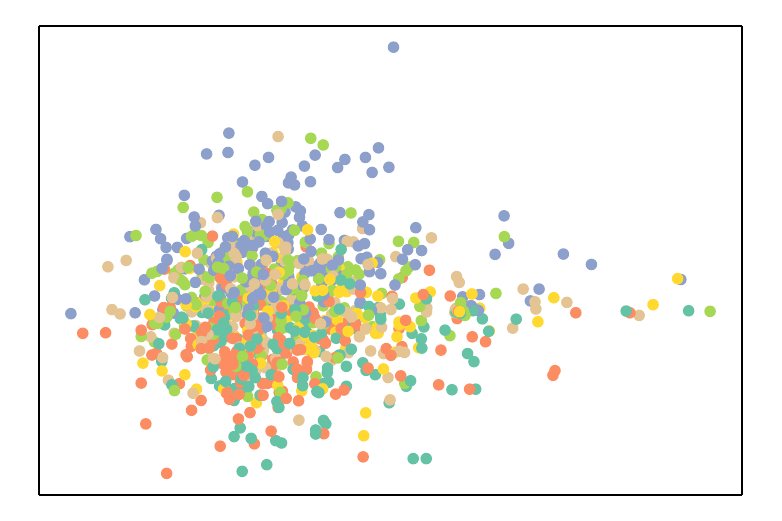}
\hspace{0.5em}
\includegraphics[width=0.31\columnwidth]{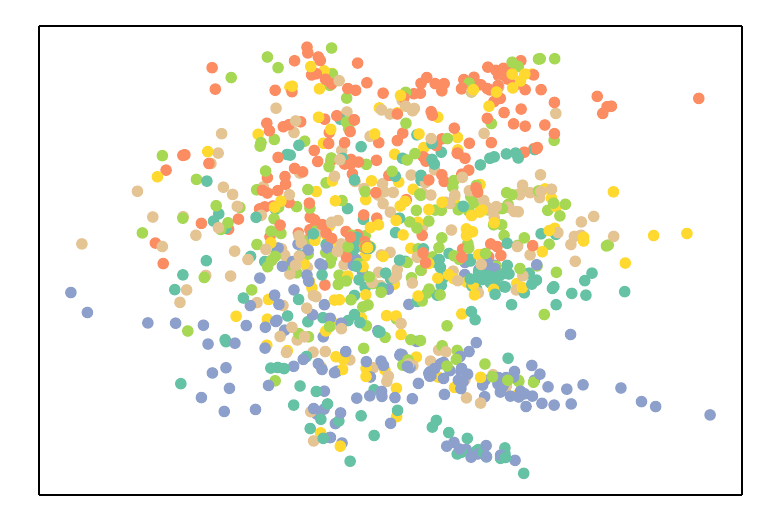}
\hspace{0.5em}
\includegraphics[width=0.31\columnwidth]{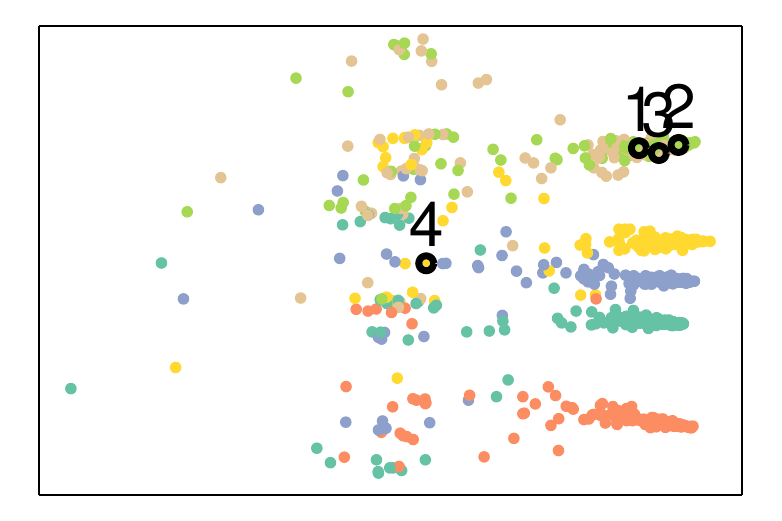}
}
\vspace{-0.5em}
\subfloat[Hard Clustering, 2048 Nodes]{
\includegraphics[width=0.31\columnwidth]{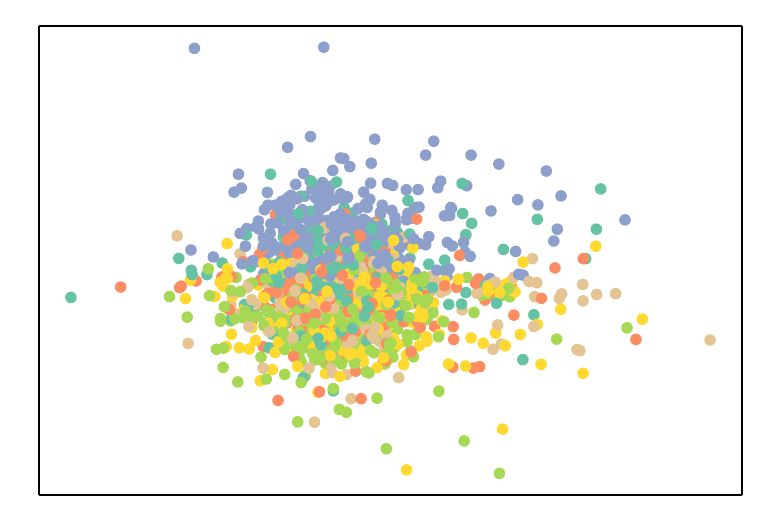}
\hspace{0.5em}
\includegraphics[width=0.31\columnwidth]{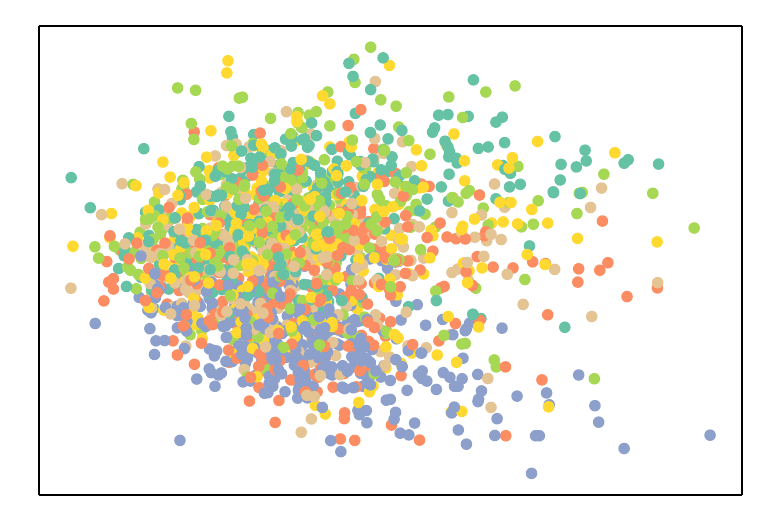}
\hspace{0.5em}
\includegraphics[width=0.31\columnwidth]{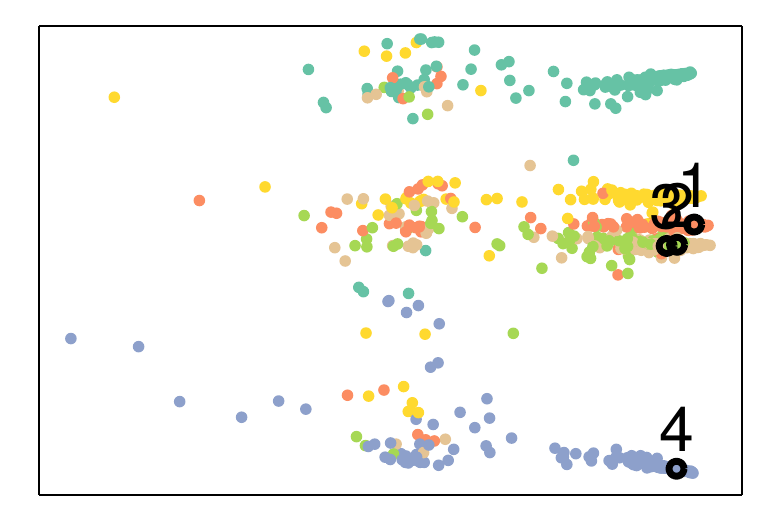}
}
\vspace{-0.5em}
\subfloat[Hard Clustering, 4096 Nodes]{
\includegraphics[width=0.31\columnwidth]{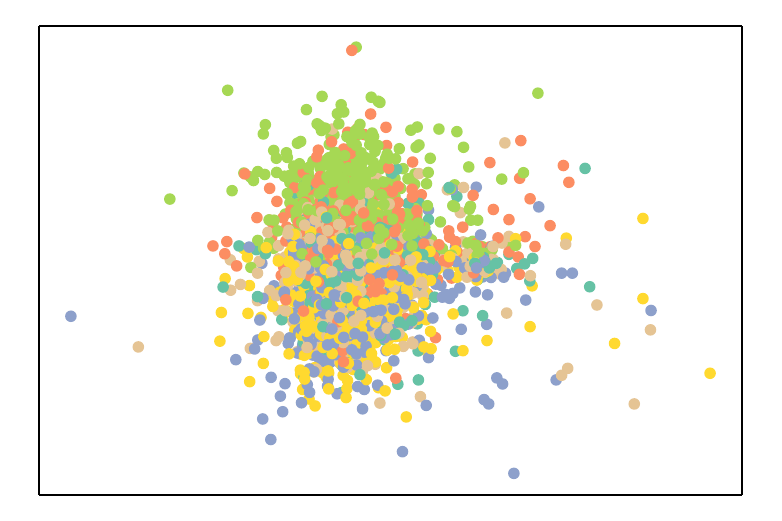}
\hspace{0.5em}
\includegraphics[width=0.31\columnwidth]{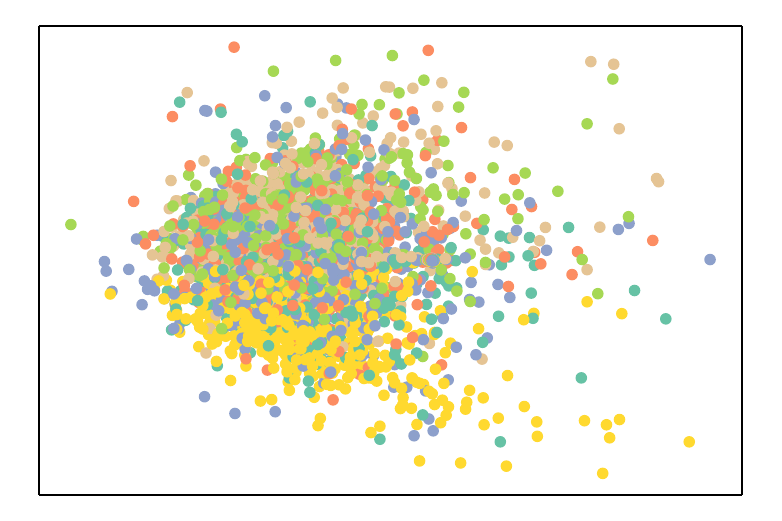}
\hspace{0.5em}
\includegraphics[width=0.31\columnwidth]{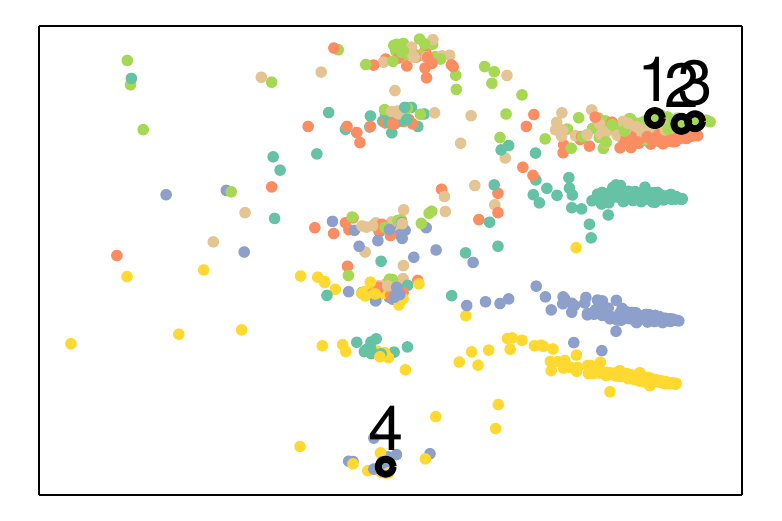}
}
\captionsetup[subfigure]{labelformat=empty}
\hspace*{\fill}
\subfloat[1. Drum]{
\includegraphics[width=0.2\columnwidth]{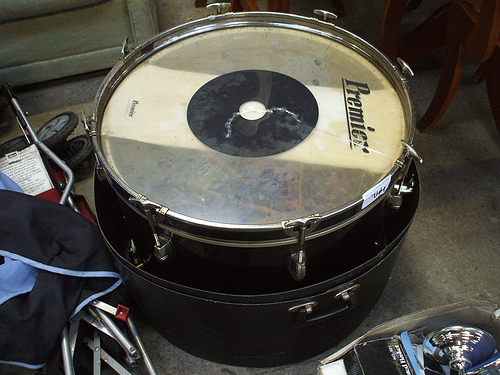}
}
\hspace*{\fill}
\subfloat[2. Great White]{
\includegraphics[width=0.2\columnwidth]{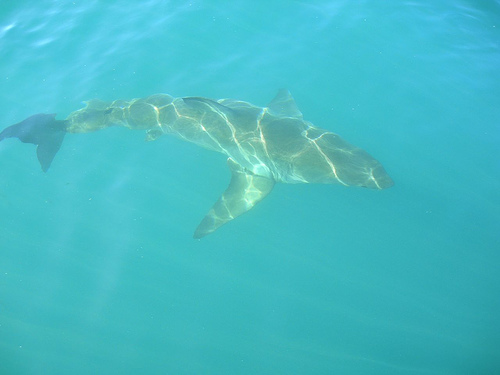}
}
\hspace*{\fill}
\subfloat[3. Tiger Shark]{
\includegraphics[width=0.2\columnwidth]{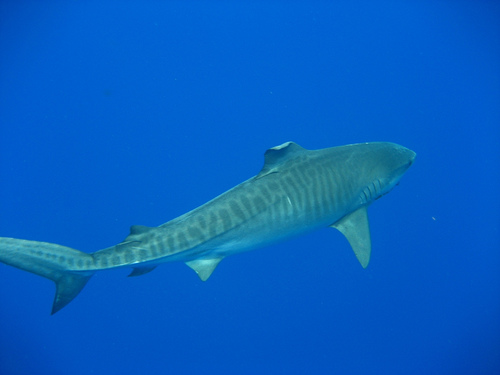}
}
\hspace*{\fill}
\subfloat[4. Hammerhead]{
\includegraphics[width=0.2\columnwidth]{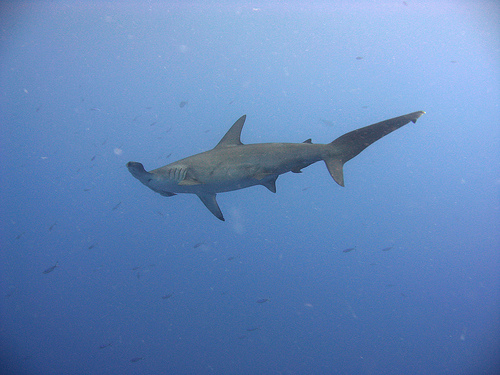}
}
\hspace*{\fill}
\end{minipage}
\caption{ (Top) A visualization of the node cluster probabilities $\mathbf{P}_j$ projected to two dimensions using PCA for models with (a) 1024 nodes, (b) 2048 nodes, and (c) 4096 nodes in each Blockout layer. Dots indicate nodes while color indicates the cluster with the highest probability. Nodes that are located close together share similar cluster assignments. Some example output categories (1-4) are also shown. (Bottom) The associated categories along with sample images. Despite the somewhat non-intuitive structure, there are clear, consistent groupings of nodes, especially in later layers and with fewer nodes.  }
\label{fig:imagenet_projection}
\end{figure}

Figure~\ref{fig:imagenet_projection} visualizes the $k$-dimensional cluster probability vectors for each node by projecting them to two dimensions using PCA. Because nodes can belong to multiple clusters with varying relative frequencies, the cluster probabilities can be interpreted as embeddings where nodes with similar probabilities indicate computations following similar paths in the hierarchical architecture. Again note that earlier layers tend to be less separated, especially with higher network capacity,  allowing for more information to be shared between clusters. In addition, because this implicit node embedding is a side effect of maximizing prediction accuracy, the resulting clusters are less interpretable than an explicit clustering based on category or image similarity. For example, while nearly indistinguishable classes such as ``great white shark'' and ``tiger shark'' do share very similar cluster probabilities, so does the visually dissimilar and seemingly unrelated class ``drum.'' Furthermore, while one might expect ``hammerhead shark'' to be close to the other sharks, it actually belongs to a completely different set of clusters. Despite this, the final cluster probabilities do seem to be fairly consistent across different choices of hyper-parameters. This could indicate that the node clusters are indeed a function of the training data, but just incorporate more information than just visual similarity, e.g. cluster balance, classification difficulty, etc. 

\begin{figure}[t!]
\centering
\subfloat[Hard Clustering, 8192 Nodes]{
\includegraphics[width=.5\columnwidth]{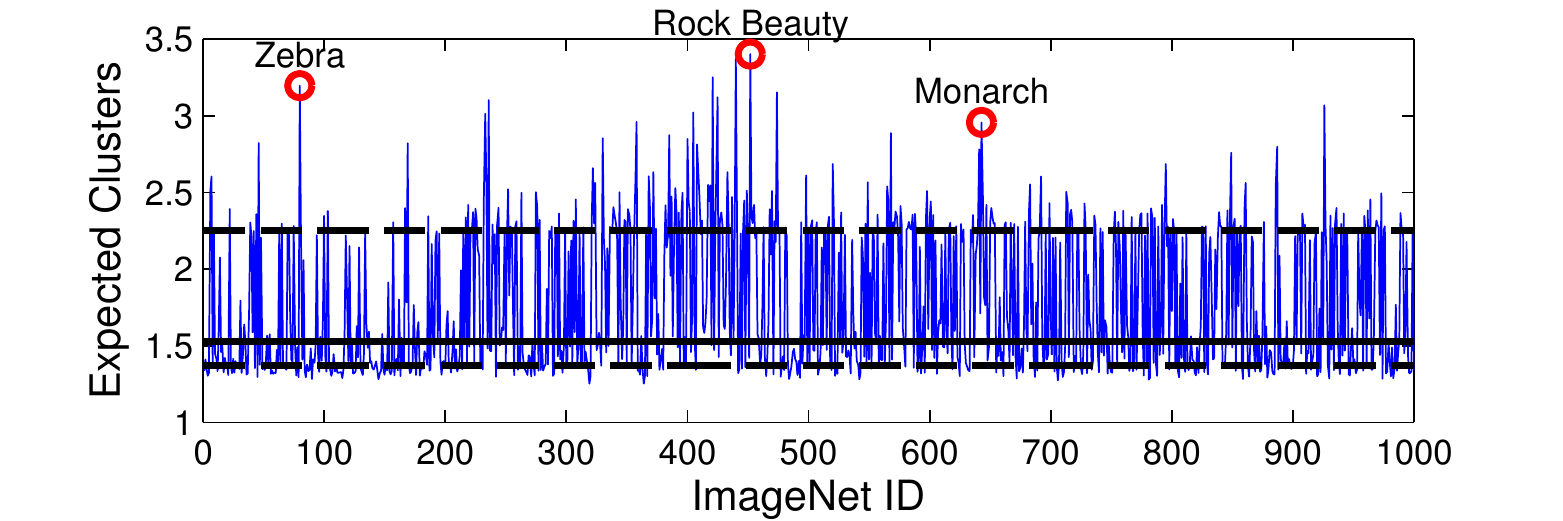}
}\hspace{-2.5em}
\subfloat[Hard Clustering, 4096 Nodes]{
\includegraphics[width=.5\columnwidth]{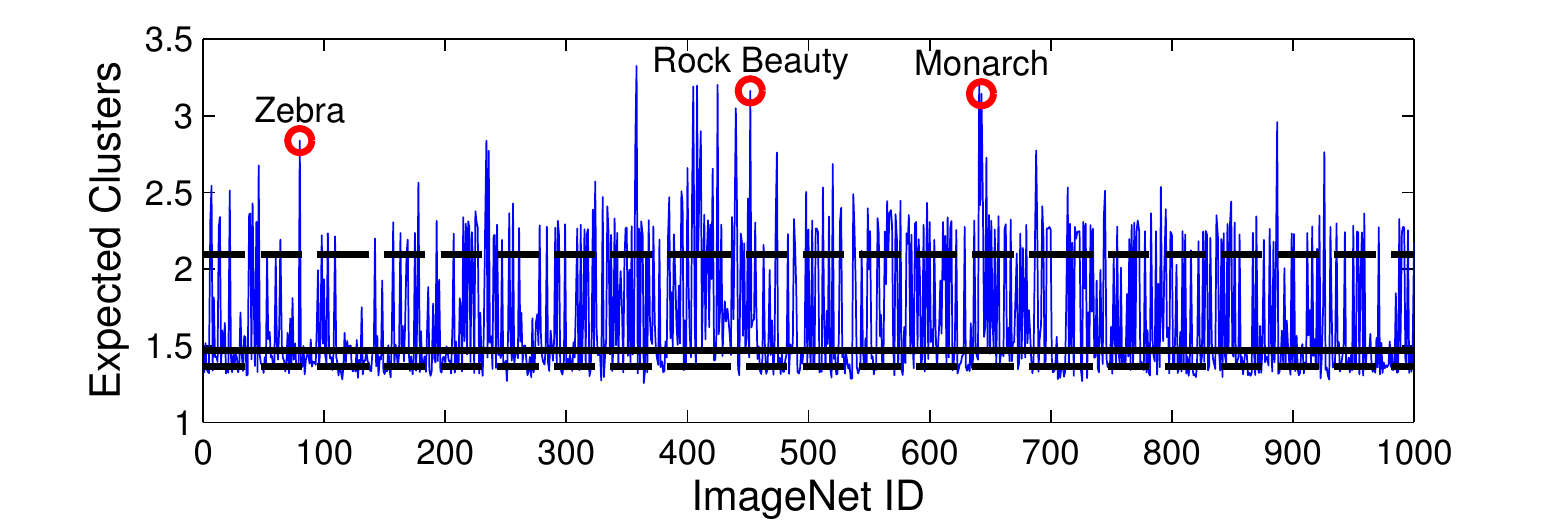}
}
\vspace{-0.5em}
\subfloat[Hard Clustering, 2048 Nodes]{
\includegraphics[width=.5\columnwidth]{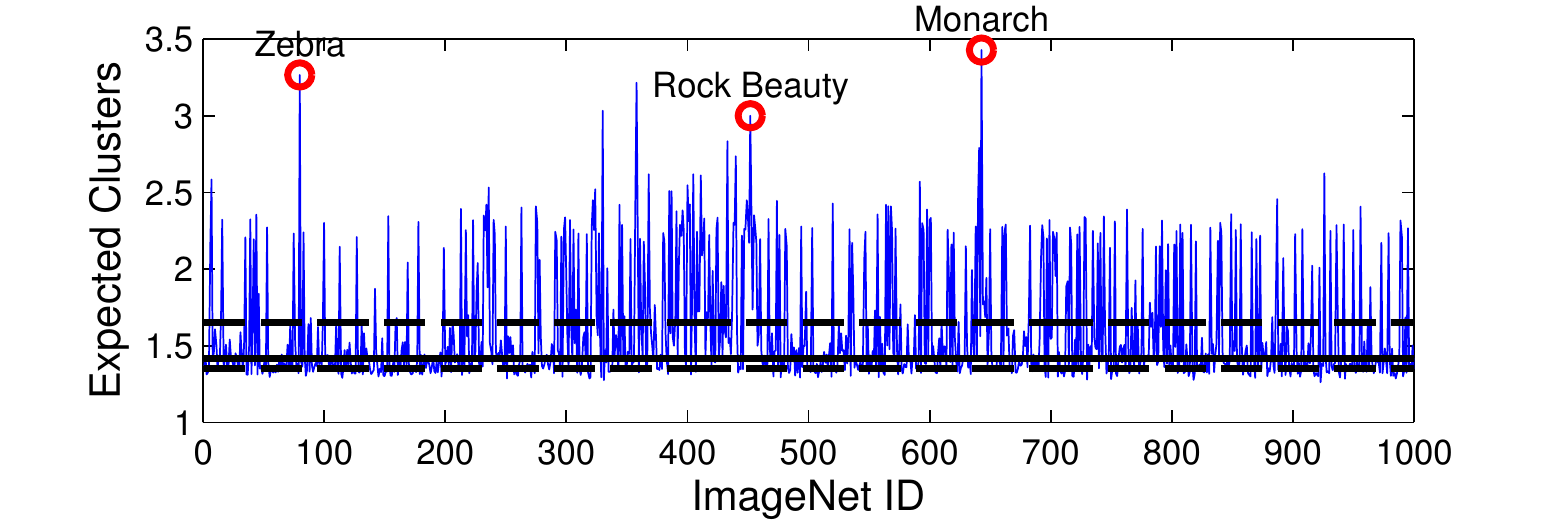}
}\hspace{-2.5em}
\subfloat[Hard Clustering, 1024 Nodes]{
\includegraphics[width=.5\columnwidth]{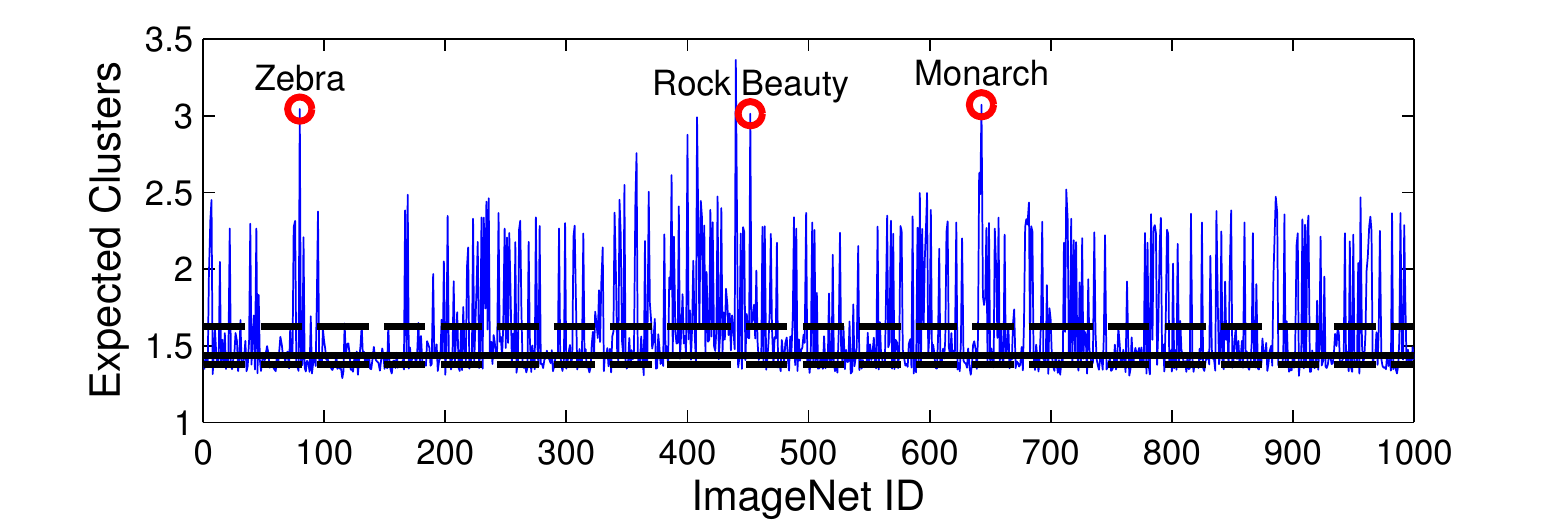}
}

\begin{minipage}{1\columnwidth}
\centering
\captionsetup[subfigure]{labelformat=empty}
\subfloat[Zebra]{
\includegraphics[width=0.15\columnwidth]{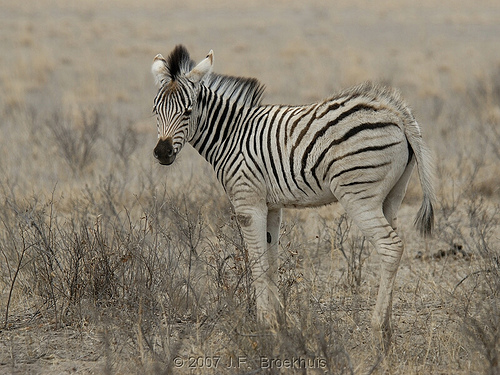}
\includegraphics[width=0.15\columnwidth]{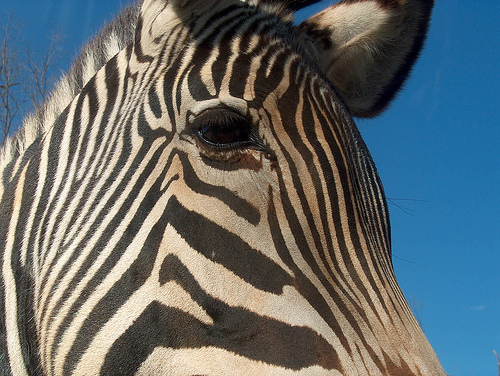}
}\hspace*{\fill}
\subfloat[Rock Beauty]{
\includegraphics[width=0.15\columnwidth]{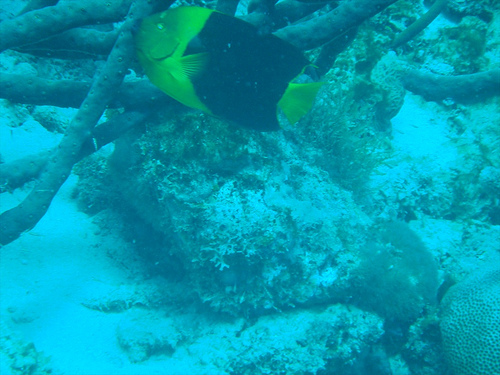}
\includegraphics[width=0.15\columnwidth]{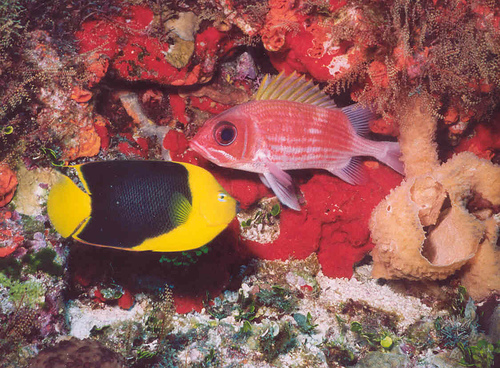}
}\hspace*{\fill}
\subfloat[Monarch]{
\includegraphics[width=0.15\columnwidth]{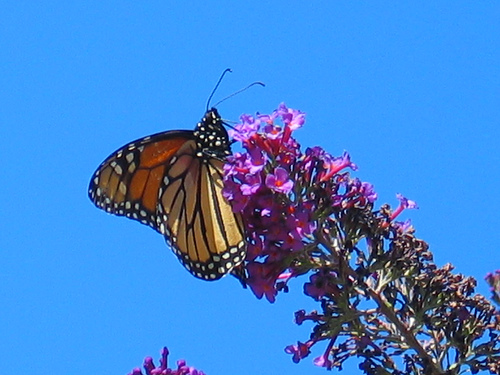}
\includegraphics[width=0.15\columnwidth]{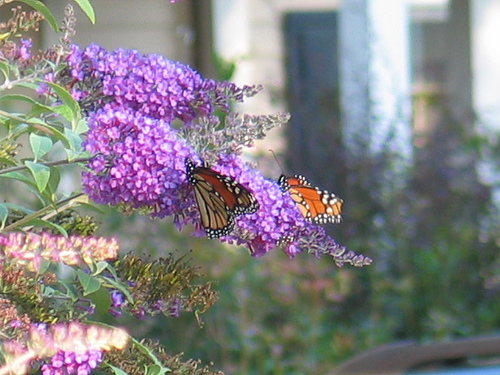}
}
\end{minipage}
\caption{ (Top) The expected number of clusters assigned to each of the ImageNet output categories for (a) 8192 nodes, (b) 4096 nodes, (c) 2048 nodes, and (d) 1024 nodes in each Blockout layer. The solid black line shows the median number of clusters while the dotted black lines show the 25th and 75th percentiles. Also shown are 3 example categories with a relatively high expected number of clusters. (Bottom) Sample images from the indicated categories, showing classification challenges such as camouflage, varied background appearance, and small relative size.}
\label{fig:imagenet_difficulty}
\end{figure}

Figure~\ref{fig:imagenet_difficulty} shows the expected number of clusters assigned to each output category. Again, notice the consistency across different hyper-parameter selections. While the median number of clusters is around 1.5, some categories belong to close to 3 with many more parameters used in their predictions. These could correspond to categories that are more difficult to predict, perhaps due to natural camouflage (e.g. ``zebra''), large variations in background appearance, or the small relative size of the subject (e.g. ``rock beauty'' and ``monarch butterfly'').

\footnotetext{This is the accuracy of our implementation of the model in~\cite{ioffe2015batch}, which reported a maximum accuracy of 74.8\%. This discrepancy is most likely due to a different learning rate decay schedule.}

\subsection{CIFAR-10}

Both CIFAR-100 and ImageNet contain a large number of categories with clear hierarchical relationships that can be  leveraged by Blockout. We also compare our approach on tasks for which these hierarchical assumptions do not necessarily hold. CIFAR-10 is a version of the CIFAR-100 dataset that classifies images into only 10 classes instead of 100. With such a small number of categories, the utility of hierarchical architectures is less clear. 
Table~\ref{tab:cifar10_acc} shows the performance of our model with only 2 clusters in comparison to the baseline, the baseline with fully-connected layers, and the baseline with fully-connected layers followed by 50\% Dropout. 
Note that stochastic regularization is less important (though still gives better performance than Dropout) while just using the soft clustering parametrization achieved the best performance of 94.41\%. This accentuates the regularization effect of the shared cluster assignments between adjacent blockout layers, even without stochastic regularization.

\section{Conclusion}

Blockout is a novel generalization of stochastic regularization with parameters that can be learned during training, essentially allowing for automatic model selection within a class of hierarchical network structures. The result is a consistent, implicit clustering of output categories into branches sharing similar representations. While further work is required to completely understand and interpret the learned clusters, Blockout results in substantial improvements in prediction accuracy and faster convergence in comparison to baseline methods. As a first step towards fully-automatic model selection, Blockout emphasizes the importance of the careful parametrization of deep network architectures and should inspire a family of similar approaches adapted to other application domains.

{
\bibliographystyle{ieee}
\bibliography{refs}
}

\end{document}